\documentclass[12pt]{article}
\usepackage{amsmath}
\usepackage{times}
\usepackage{graphicx}
\usepackage{color}
\usepackage{multirow}
\usepackage[authoryear]{natbib}
\usepackage{rotating}
\usepackage{bbm}
\usepackage{latexsym}

\usepackage{subfigure}
\usepackage{algorithm}
\usepackage{algorithmic}
\usepackage{hyperref}

\usepackage{amssymb}
\usepackage{eucal}
\usepackage{amsthm}
\usepackage{nccbbb}
\usepackage{tabularx}
\usepackage{booktabs} 
\usepackage{galois} 
\usepackage{sidecap}
\usepackage{multirow}

\newcommand{\captionfonts}{\normalsize}

\makeatletter
\long\def\@makecaption#1#2{%
  \vskip\abovecaptionskip
  \sbox\@tempboxa{{\captionfonts #1: #2}}%
  \ifdim \wd\@tempboxa >\hsize
    {\captionfonts #1: #2\par}
  \else
    \hbox to\hsize{\hfil\box\@tempboxa\hfil}%
  \fi
  \vskip\belowcaptionskip}
\makeatother

\def\X{\mathbf{x}}
\def\Y{\mathbf{y}}
\def\Z{\mathbf{z}}
\def\W{\boldsymbol{\omega}}
\def\Wt{\W^{\prime}}
\def\I{\mathbf{I}}
\def\E{\mathbb{E}}
\def\M{\CMcal{M}}

\begin{document}
\hspace{13.9cm}1

\ \vspace{20mm}\\

{\LARGE FastMMD: Ensemble of Circular Discrepancy for Efficient Two-Sample Test}

\ \\
{\bf \large Ji Zhao$^{\displaystyle 1}$, Deyu Meng$^{\displaystyle 2}$}\\
{$^{\displaystyle 1}$\url{zhaoji84@gmail.com} \\ The Robotics Institute, Carnegie Mellon University, PA 15213, USA.}\\
{$^{\displaystyle 2}$\url{dymeng@mail.xjtu.edu.cn} \\ Xi'an Jiaotong University, Xi'an 710049, China.}\\
%


\thispagestyle{empty}
\markboth{}{NC instructions}
\ \vspace{-0mm}\\
%
\begin{center} {\bf Abstract} \end{center}
The maximum mean discrepancy (MMD) is a recently proposed test statistic for two-sample test. Its quadratic time complexity, however, greatly hampers its availability to large-scale applications. To accelerate the MMD calculation, in this study we propose an efficient method called FastMMD. The core idea of FastMMD is to equivalently transform the MMD with shift-invariant kernels into the amplitude expectation of a linear combination of sinusoid components based on Bochner's theorem and Fourier transform \citep{Rahimi07}. Taking advantage of sampling of Fourier transform, FastMMD decreases the time complexity for MMD calculation from $O(N^2 d)$ to $O(L N d)$, where $N$ and $d$ are the size and dimension of the sample set, respectively. Here $L$ is the number of basis functions for approximating kernels which determines the approximation accuracy. For kernels that are spherically invariant, the computation can be further accelerated to $O(L N \log d)$ by using the Fastfood technique \citep{LeQ13}. The uniform convergence of our method has also been theoretically proved in both unbiased and biased estimates. We have further provided a geometric explanation for our method, namely ensemble of circular discrepancy, which facilitates us to understand the insight of MMD, and is hopeful to help arouse more extensive metrics for assessing two-sample test. Experimental results substantiate that FastMMD is with similar accuracy as exact MMD, while with faster computation speed and lower variance than the existing MMD approximation methods.

\section{Introduction}
The two-sample test is one of the most fundamental tests in statistics and has a wide range of applications. It uses samples drawn from two distributions to test whether to accept or reject the null hypothesis that they are the same or different.
This task, however, is very difficult and challenging in practice since the underneath distribution information are generally unknown apriori \citep{Bickel69, Friedman79, Hall02, Biau05}.
The maximum mean discrepancy (MMD) is the latest test statistic designed for this task by measuring the discrepancy of two distributions by embedding them in a reproducing kernel Hilbert space \citep{Gretton12}.
The MMD has been attracting much attention in recent two-sample test research due to its solid theoretical fundament \citep{Smola07, Sriperumbudur10, Sriperumbudur11, Sejdinovic13} and successful applications including biological data test, data integration and attribute matching \citep{Gretton09}, outlier detection, data classifiability \citep{Sriperumbudur09}, domain adaption, etc.
By generalizing the MMD to kernel families as the supremum of MMDs on a class of kernels, it has also been effectively used for some basic machine learning problems such as kernel selection \citep{Sriperumbudur09}.

Albeit its various applications, the exact MMD needs $O(N^2 d)$ computational cost, where $N$ and $d$ denote the size and dimension of samples, respectively, to calculate the kernel values between all pairs from the assessed two-sample sets. This quadratic computational complexity greatly hampers its further application to large-scale practical problems. How to speedup the computation of MMD has thus become a hot issue in statistics and machine learning in recent years.

There are mainly two approaches proposed for this problem by approximating MMD on a subsampling set of all sample pairs. The first is \emph{MMD-linear}, which is the extremely simplified MMD calculation by only using possibly fewest interactions of sample pairs \citep{Gretton12}. While this strategy significantly accelerates the MMD calculation to $O(N d)$, it also brings very high variance due to its evident loss of sample pair information. To better leverage the computation cost and calculation accuracy of MMD, B-test is recently proposed \citep{Zaremba13}. The main idea is to split two-sample sets into corresponding subsets, construct block correspondence between them, and then compute the exact MMD inner each block while omit the inter-block pair information. By changing the block size, it can vary smoothly from \emph{MMD-linear} with linear complexity to exact MMD with quadratic complexity. In practice, the block size is generally set as a modest value $\sqrt{N}$ by experience. Thus the time complexity of \emph{B}-test is $O(N^{3/2} d)$, correspondingly.

Actually, as the coming of the big data era, it has become a hot trend to enhancing the efficiency of kernel-based learning methods, such as support vector machines and Gaussian process, throughout machine learning, computer vision and data mining.
Many efforts have been made to speedup the establishment of the kernel information and accelerate the implementation of kernel techniques \citep{Smola00, Williams00, Fine01}. Two of the representative developments are Random Kitchen Sinks \citep{Rahimi07, Rahimi08} and Fastfood \citep{LeQ13}, which can significantly speed up the computation for a large range of kernel functions by mapping data into a relatively low-dimensional randomized feature space. These developments inspire us for this MMD-acceleration research topic, which constitutes an important branch along this line of research.

The main difficulty and challenge of MMD calculation lie in the fact that it needs to compute the kernel values between all sample pairs of two sets. \emph{MMD-linear} and \emph{B}-test attain this task by only utilizing a subsampling pair subset from all. Such simplification, however, also decreases the accuracy of MMD calculation due to their neglectness of the entire sample pair information. To this aim, this paper proposes a new efficient MMD calculation strategy, which can implement this task in a more efficient and accurate way. In summary, this paper mainly contains the following four-fold contributions:

1. Through employing Bochner's theorem and Fastfood technique \citep{LeQ13} for kernels that are spherically invariant, we reduce the MMD computation cost to $O(L N \log d)$, which has lower time complexity than current MMD approximation methods
and facilitates MMD's application in large-scale data. Moreover, our method is easy to be sequentially computed and parallelized.

2. The proposed method utilizes the interacted kernel values between all pairs from two sample sets to calculate MMD, which naturally leads to very accurate MMD result. Our experimental results substantiate that our method is with similar accuracy as exact MMD, and with significantly smaller variance than \emph{MMD-linear} and \emph{B}-test.

3. We have theoretically proved the uniform convergence of our method in both unbiased and biased cases. Comparatively, both \emph{MMD-linear} and \emph{B}-test are only feasible in unbiased cases.

4. We provide a geometrical explanation of our method in calculating MMD with shift-invariant kernels. Under this viewpoint, it is potentially useful for arousing more extensive metrics for two-sample test.

The code of our FastMMD method is available at \url{http://gr.xjtu.edu.cn/web/dymeng/2}.

\section{Efficient MMD for Shift-Invariant Kernels}
Firstly we give a brief review of MMD \citep{Gretton12} and introduce some important properties of shift-invariant kernel \citep{Rahimi07}. Then we propose an efficient MMD approximation method.

Consider a set of samples drawn from two \mbox{distributions}
$S = \{ ( \X_i, \ell_i ) \in {\bbbr}^d \times \{1,2\} \}^{N}_{i=1}$,
where the label $\ell_i$ indicates the distribution from which $\X_i$ is drawn.
The indices of samples with label $\{1, 2\}$ are denoted by $I_{1} = \{ i \, | \, \ell_i = 1 \}$ and $I_{2} = \{ i \, | \, \ell_i = 2 \}$, respectively.

\subsection{Overview of MMD}
\noindent {\bf Definition~1} \ \ \emph{
Let $p_1(\X)$, $p_2(\X)$ be distributions defined on a domain ${\bbbr}^d$. Given observations $\{ ( \X_i, \ell_i ) \}_{i=1}^N$, where $X_1 = \{ \X_i | \ell_i = 1 \}$ and $X_2 = \{ \X_i | \ell_i = 2 \}$ are i.i.d. drawn
from $p_1(\X)$ and $p_2(\X)$, respectively.
Denote $I_{1} = \{ i \, | \, \ell_i = 1 \}$ and $I_{2} = \{ i \, | \, \ell_i = 2 \}$.
Let $\mathcal{F}$ be a class of functions $f: {\bbbr}^d \rightarrow {\bbbr}$. Then the maximum mean discrepancy and its empirical estimate are defined as (Definition~2 in \citet{Gretton12})\footnote{The empirical MMD is dependent on two compared sample sets $X_1$ and $X_2$. In the following text, we omit these two terms for notion convenience in some cases.}:
\begin{align*}
& \text{\rm MMD}[\mathcal{F}, p_1, p_2]  =  \sup_{f \in \mathcal{F}} \left( \E_{\X\sim p_1} f(\X) - \E_{\X\sim p_2} f(\X) \right), \\
& \text{\rm MMD}[\mathcal{F}, X_1, X_2]  =  \sup_{f \in \mathcal{F}} \left( \frac{1}{|I_1|}\sum_{i \in I_1} f(\X_i) - \frac{1}{|I_2|}\sum_{i \in I_2} f(\X_i) \right).
\end{align*}
}

Usually, $\mathcal{F}$ is selected to be a unit ball in a characteristic RKHS $\mathcal{H}$, defined on the metric space ${\bbbr}^d$ with associated kernel $K(\cdot, \cdot)$ and feature mapping $\phi(\cdot)$. The popular Gaussian and Laplacian kernels are characteristic \citep{Sriperumbudur11}. If $\int \sqrt{K(\X, \X)} \mathrm{d} p_1(\X) < \infty$ and $\int \sqrt{K(\X, \X)} \mathrm{d} p_2(\X) < \infty$, we denote $\mu(p) = \E_{\X\sim p(\X)}\phi(\X)$ as the expectation of $\phi(\X)$. Then it has been proved that (Lemma 4 in \citet{Gretton12}):
\begin{equation*}
\text{MMD}[\mathcal{F}, p_1, p_2]  = \left\| \mu(p_1) - \mu(p_2) \right\|_{\mathcal{H}}.
\end{equation*}
Substituting the empirical estimates $\mu(X_1):= \frac{1}{|I_1|}\sum_{i \in I_1} \phi(\X_i)$ and $\mu(X_2):= \frac{1}{|I_2|}\sum_{i \in I_2} \phi(\X_i)$ of the feature space means based on respective samples, an empirical biased estimate of MMD can then be obtained as:
\begin{align}
\text{\rm MMD}_{\text{b}}[\mathcal{F}, X_1, X_2]  &= \left\| \sum_{i=1}^N a_i \phi(\X_i) \right\|_{\mathcal{H}}  =  \left[ \sum_{i=1}^N \sum_{j=1}^N a_i a_j K(\X_i, \X_j) \right]^{\frac{1}{2}},
\end{align}
where $a_i = \frac{1}{|I_1|}$ if $i \in I_1$, and $a_i = \frac{-1}{|I_2|}$ if $i \in I_2$.
We can see that the time complexity of such MMD estimate is $O(N^2 d)$. We will investigate how to
accelerate its computation to $O(L N \log d)$, especially for shift-invariant kernels.

\subsection{Efficient Approximation of MMD}
The following classical theorem from harmonic analysis provides the main fundament underlying our approximation method \citep{Genton01}.

\noindent {\bf Theorem~1 (Bochner)} \ \ \emph{
Every bounded continuous positive definite function is Fourier transform of a non-negative finite Borel measure.
This means that for any bounded shift-invariant kernel $K(\X, \Y)$, there exists a non-negative finite Borel measure $\mu$ satisfying
\begin{align*}
K(\X, \Y) = \int_{{\bbbr}^d}\! e^{j \Wt (\X - \Y)} \, \mathrm{d} \mu(\W),
\end{align*}
where $\mu(\W)$ is Fourier transform of kernel $K(\boldsymbol{\Delta})$, and its normalization $p(\W) = \mu(\W) / \int \mathrm{d} \mu(\W)$ is a probability measure. Here $j = \sqrt{-1}$ is the imaginary unit.
}

We assume that the discussed positive definite kernel is real valued.
According to Bochner's theorem, we have that if a shift-invariant kernel $K(\X, \Y)$ is positive definite, there exists a proper scaled probability measure $p(\W)$ satisfying
\begin{align}
K(\X, \Y) = & K(\mathbf{0}) \cdot \int_{{\bbbr}^d}\!  e^{j \Wt (\X - \Y)} \, \mathrm{d} p(\W) \nonumber \\
= & K(\mathbf{0}) \cdot \int_{{\bbbr}^d}\!  \cos(\Wt \X - \Wt \Y) \, \mathrm{d} p(\W). \label{equ:fmap}
\end{align}
Since both the probability measure $p(\W)$ and the kernel $K(\Delta)$ are real, the integrand $e^{j \Wt (\X - \Y)}$ can be replaced by $\cos(\Wt \X - \Wt \Y)$ in the above equation.
Taking the Gaussian kernel $K(\X, \Y; \sigma) = e^{- \frac{\| \X - \Y \| ^2}{2 {\sigma}^2}}$ as an example, we can rewrite it as $K(\boldsymbol{\Delta}; \sigma) = e^{- \frac{\| \boldsymbol{\Delta} \| ^2}{2 {\sigma}^2}}$, where $\boldsymbol{\Delta} = \X-\Y$. Its Fourier transform is $p(\W; \sigma) = (2 \pi)^{- \frac{d}{2}} e^{- \frac{\sigma^2 \| \W \| ^2}{2} }$. By proper scaling, $\W$ can be viewed as a multivariate Gaussian distribution $\W \sim \CMcal{N} (\mathbf{0}, \frac{1}{\sigma^2} \I)$, where $\I$ is the $d \times d$ identity matrix.

\noindent {\bf Claim~2} \ \ \emph{
For a shift-invariant kernel  $K(\X,\Y)=\left\langle\phi(\X),\phi(\Y)\right\rangle$,
suppose $p(\W)$ is its corresponding normalized measure in Bochner's theorem, then
\begin{align}
& \sum_{i=1}^N \sum_{j=1}^N a_i a_j K(\X_i, \X_j) = K(\mathbf{0}) \cdot \E_{\W\sim p(\W)} \sum_{i=1}^N \sum_{j=1}^N a_i a_j \cos(\Wt\X_i-\Wt\X_j). \label{equ:equv}
\end{align}
}

\begin{figure}[!htbp]
    \centering
    \includegraphics[width=0.98\linewidth]{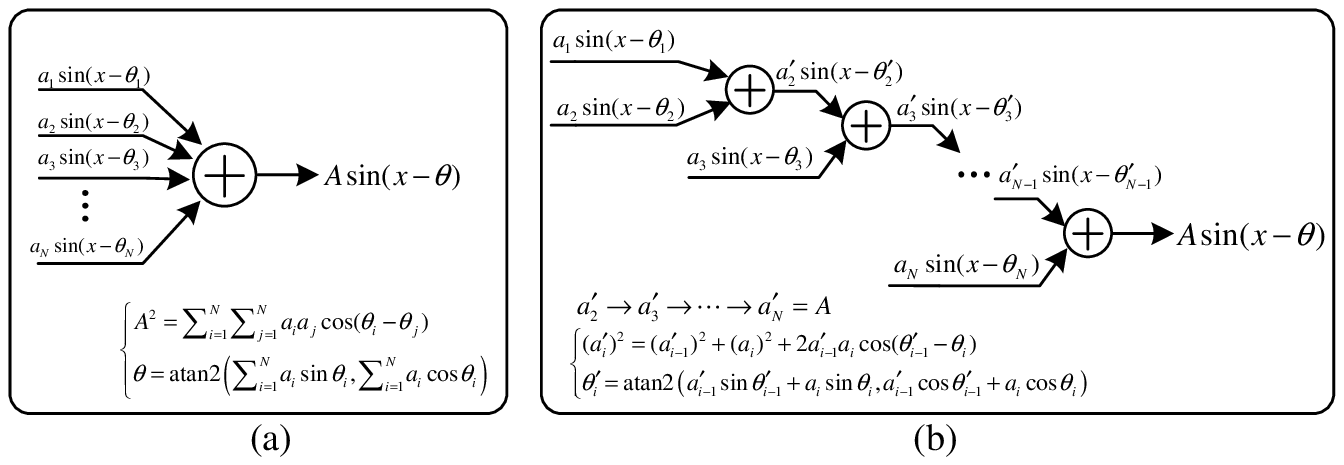}
\caption{Amplitude of $N$ combined sinusoids (a) by direct trigonometric summarization and (b) by sequential summarization.}
\label{fig:sinusoid}
\end{figure}

Claim~2 can be easily proved by substituting Eqn.~\eqref{equ:fmap} into kernel $K(\X_i, \X_j)$.
A very interesting thing is that we can fortunately calculate
$\sum_{i=1}^N \sum_{j=1}^N a_i a_j \cos(\Wt\X_i-\Wt\X_j)$ in linear time by applying the Harmonic Addition Theorem \citep{Nahin95}, as shown in Fig.~\ref{fig:sinusoid}.
First, this expression can be viewed as a squared amplitude of combined sinusoids.
Suppose a linear combination of $N$ sinusoids is $\sum_{i=1}^N a_i \sin(x - \Wt\X_i) = A \sin(x - \theta)$, then its amplitude has a closed form $A^2 = \sum_{i=1}^N \sum_{j=1}^N a_i a_j \cos(\Wt\X_i-\Wt\X_j)$, see Fig.~\ref{fig:sinusoid}(a).
Second, the amplitude of sinusoids with the same frequency can be calculated in a sequential way in linear time, see Fig.~\ref{fig:sinusoid}(b).
By combining the above two observations, we can calculate the expression with linear time complexity.
If we set $a_i$ in Eqn.~\eqref{equ:equv} as that in empirical MMD, it turns out to be the biased estimate of MMD. As a result, Claim~2 finely implies a novel methodology to efficiently approximate MMD.

By the same spirit of Random Fourier features \citep{Rahimi07}, we draw i.i.d samples from distribution $p(\W)$, and then take the average of $A^2(\W; X_1, X_2)$ to approximate the empirical estimate of squared MMD. The Quasi-Monte Carlo sampling techniques can also be used to generate sample sequence which has lower discrepancy than i.i.d sampling \citep{YangJ14}.
Formally, we sample $L$ samples $\{ \W_k \}_{k=1}^L$ from the distribution $p(\W)$, and then use them to approximate $\text{\rm MMD}_{\text{b}}$:
\begin{equation}
\overline{\text{\rm MMD}}_{\text{b}} = \left[ \frac{K(\mathbf{0})}{L} \sum_{k=1}^{L} A^2 (\W_k; X_1, X_2) \right]^{\frac{1}{2}},
\end{equation}
where $A(\W_k; X_1, X_2)$ is the amplitude of linear combination of sinusoids $\frac{1}{|I_1|} \sum_{i \in I_1} \sin(x - \Wt\X_i) - \frac{1}{|I_2|} \sum_{i \in I_2}  \sin(x - \Wt\X_i)$.
We call $L$ the number of basis functions \citep{Neal94, LeQ13}.
The efficient calculation of $\overline{\text{\rm MMD}}_{\text{u}}$ for unbiased MMD is similar since we can also rewrite unbiased estimate of MMD as the form in Eqn.~\eqref{equ:equv}. The proof is provided in Appendix~A.1. The aforementioned procedure for approximating MMD is described in Algorithm~\ref{alg:alg1}. We also provide an equivalent implementation of Algorithm~\ref{alg:alg1} in Appendix~A.2.

\begin{algorithm}[!htbp]
   \caption{FastMMD for shift-invariant kernels} \label{alg:alg1}
   \textbf{Input:} Sample set $S = \{ ( \X_i, \ell_i ) \}^{N}_{i=1}$; shift-invariant kernel $K(\boldsymbol{\Delta})$. Denote $I_{1} = \{ i \, | \, \ell_i = 1 \}$, $I_{2} = \{ i \, | \, \ell_i = 2 \}$ \\
   \textbf{Output:} MMD approximation $\overline{\text{\rm MMD}}_{\text{b}}^2$, $\overline{\text{\rm MMD}}_{\text{u}}^2$.
\begin{algorithmic}[1]
   \STATE Calculate Fourier transform $\mu(\W)$ of $K(\boldsymbol{\Delta})$, and set $p(\W) = \mu(\W)/K(\mathbf{0})$.
   \STATE Calculate $\{\Wt_k\X_i\}$, where $\{\W_k\}^{L}_{k=1}$ are samples drawn from $p(\W)$.
   \FOR{$k=1$ {\bfseries to} $L$}
   \STATE Calculate amplitude $A_1(\W_k)$ and phase $\theta_1(\W_k)$ for $\frac{1}{|I_1|} \sum_{i \in I_1} \sin(x - \Wt_k\X_i)$.
   \STATE Calculate amplitude $A_2(\W_k)$ and phase $\theta_2(\W_k)$ for $\frac{1}{|I_2|} \sum_{i \in I_2} \sin(x - \Wt_k\X_i)$.
   \STATE $A^2(\W_k) = A_1^2(\W_k) + A_2^2(\W_k) - 2 A_1(\W_k) A_2(\W_k) \cos(\theta_1(\W_k) - \theta_2(\W_k))$.
   \ENDFOR
   \STATE $\overline{\text{\rm MMD}}_{\text{b}}^2 = \frac{K(\mathbf{0})}{L} \sum_{k=1}^{L} A^2(\W_k)$.
   \STATE $\overline{\text{\rm MMD}}_{\text{u}}^2 = \frac{K(\mathbf{0})}{L} \big[ \sum_{k=1}^{L} A^2(\W_k) + \frac{1}{|I_1|-1}  \sum_{k=1}^{L} A_1^2(\W_k)$ \\ $\qquad \qquad + \frac{1}{|I_2|-1}  \sum_{k=1}^{L} A_2^2(\W_k) \big] - \frac{|I_1|+|I_2|-2}{(|I_1|-1)(|I_2|-1)} K(\mathbf{0})$.
\end{algorithmic}
\end{algorithm}

For kernels that are spherically invariant, the Fastfood technique can be employed to further speedup $\W$ sampling in step 2 and $\Wt \X_u$ calculation in steps 4 - 5 of Algorithm~\ref{alg:alg1} \citep{LeQ13}.
This can bring further efficiency gain for MMD calculation from $O(L N d)$ to $O(L N \log d)$. In the rest of this paper, we call our original algorithm as \emph{FastMMD-Fourier} and its variant using Fastfood as  \emph{FastMMD-Fastfood}, respectively.

\subsection{Computational Complexity}
As aforementioned, $N$ is the number of samples, $d$ is the dimension number of samples, and
$L$ is the number of basis functions for approximating $p(\W)$.
Given a sampling of $\W$, the time complexity for calculating $A(\W; X_1, X_2)$ is $O(N d)$.
The overall computational complexity of the entire \emph{FastMMD-Fourier} is thus $O(LNd)$.
For \emph{FastMMD-Fastfood}, the computation speed is further enhanced to $O(L N \log d)$.
Usually, the basis number $L$ can preset as a fixed number and are thus independent of the sample scale.
As compared with the complexities of the previous MMD methods, such as $O(N^2 d)$ for exact MMD \citep{Gretton12}, 
and $O(N^{3/2}d)$ for \emph{B}-test \citep{Zaremba13}, the proposed FastMMD methods evidently get a speed gain. Furthermore, instead of only utilizing a subsampling pair subset from all by the current MMD approximation methods, FastMMD takes into consideration all the interacted information between sample pairs. Our methods are thus expected to be more accurate.

Another interesting thing is that the calculation of $A(\W; X_1, X_2)$ can be computed in a sequential way, and thus our method can be naturally implemented in stream computations. Also our method is easy to be parallelized. This further implies the potential usefulness of the proposed FastMMD methods in real large-scaled applications.

\subsection{Approximation Guarantees}
In this section, we will prove the approximation ability of the proposed FastMMD methods.

\noindent {\bf Theorem~3 (Uniform Convergence of \emph{FastMMD-Fourier})} \ \ \emph{Let $\M$ be a compact subset of ${\bbbr}^d$ with diameter $\operatorname{\text{\rm diam}}(\M)$. Then, for the biased estimate of MMD
in Algorithm~\ref{alg:alg1}, we have:
\begin{align*}
\text{\rm Pr}\left[\sup_{\X_1, \cdots, \X_N \in \M} \left| \overline{\text{\rm MMD}}_{\text{\rm b}}^2 - \text{\rm MMD}_{\text{\rm b}}^2 \right| \ge \epsilon \right]
\le 2^{12} \left( \frac{\sigma_p \operatorname{\text{\rm diam}}(\M)}{\epsilon} \right)^2 \exp\left( -\frac{L \epsilon^2}{64(d+2)} \right), \nonumber
\end{align*}
where $\sigma_p^2 = \E_p[\W^{\prime}\W]$ is the second moment of the Fourier transform of kernel $K$. This bound also holds for the approximation of unbiased MMD.
}

\noindent {\bf Theorem~4 (Uniform Convergence of \emph{FastMMD-Fastfood})} \ \ \emph{If we use Fastfood method \citep{LeQ13} to calculate $\{ \Wt_k \X_i \}_{k=1}^L$ in Algorithm~\ref{alg:alg1}, suppose the kernel is Gaussian kernel with bandwidth $\sigma$ and  $\widehat{\text{\rm MMD}}_{\text{\rm b}}$ is the biased estimate of MMD that arises from a $d \times d$ block of Fastfood \footnote{In Fastfood method, the kernel expansions are calculated by constructing several $d \times d$ blocks. For the asymptotic analysis of error bound, we can treat $L$ as $d$ by padding the data with zeros if $d < L$. This allow us to consider only one block.},
then we have:
\begin{align*}
\text{\rm Pr} \bigg[ \sup_{\X_1, \cdots, \X_N \in \M} \left| \widehat{\text{\rm MMD}}_{\text{\rm b}}^2 - \text{\rm MMD}_{\text{\rm b}}^2 \right| \ge  \epsilon \bigg]
\le 2^{20} \left( \frac{\log d \operatorname{\text{\rm diam}}(\M)}{d \sigma^2 \epsilon^2} \right)^2 \exp\left( -\frac{d \epsilon^2}{64(d+2)} \right)
\end{align*}
This bound also holds for the approximation of unbiased MMD.
}

From Theorems~3 and 4, we can see that the approximation of FastMMD is unbiased. The proof is provided in Appendix~A.2, A.3 and A.4.

\subsection{Tests Based on the Asymptotic Distribution of the Unbiased Statistic}
The principle for constructing our FastMMD approximation complies with the spirit of random kitchen and sinks \citep{Rahimi07} and Fastfood \citep{LeQ13}, both using the kernel expansion to approximate shift-invariant kernels. Specifically, we use the linear combination of Dirac-delta functions to approximate continuous function $p(\W)$, which is uniquely determined by kernel $K(\cdot, \cdot)$. It means that we implicitly introduce a new kernel which is an approximation for the original kernel in MMD calculation.

Based on the aforementioned analysis, the uniform convergence bounds and the asymptotic distribution for general kernels in \citet{Gretton12}, including the Theorem and Corollary 7 - 13, still hold for FastMMD.
Given the asymptotic distribution of the unbiased statistic $\text{\rm MMD}_{\text{\rm u}}^2$, the goal of the two-sample test is to determine whether the empirical test statistic $\text{\rm MMD}_{\text{\rm u}}^2$ is so large as to be outside the $1-\alpha$ quantile of the null distribution. In \citet{Gretton12}, two approaches based on asymptotic distributions are proposed. One method is using the bootstrap \citep{Arcones92} on aggregated data, and the other is approximating the null distribution by fitting Pearson curves to its first three moments. Both the two methods can be incorporated into FastMMD.

\section{Ensemble of Circular Discrepancy}
In the previous section, we proposed an efficient approximation for MMD. In this section, we give a geometric explanation for our methods by using random projection \citep{Blum06} on a circle and circular discrepancy. This explanation is expected to help us more insightfully understand such approximation and inspire more extensive metrics for the two-sample test other than MMD.

\subsection{Random Projection on a Unit Circle}
If $\W$ in Eqn.~\eqref{equ:fmap} is fixed, we can see that the positions of points projected on a unit circle sufficiently determine the kernel. In other words, the random variables $\Wt \X$ and $\Wt \Y$ can be wrapped around the circumference of a unit circle without changing the value of kernel function.
We first investigate the circular distribution under fixed $\W$ in the following, and later will discuss the cases when $\W$ is sampled from a multivariate distribution.

Given a fixed $\W$, we wrap two classes of samples on a unit circle separately.
The probability density functions (PDFs) of the wrapped two random variables $X_1(\W)$ and $X_2(\W)$ can be mathematically expressed as:
\begin{align}
p_1(x; \W) &= \frac{1}{|I_1|} \sum_{i \in I_1} \delta \left( x - \operatorname{mod} (\W^{\prime} \X_i, 2\pi) \right), \label{X1} \\
p_2(x; \W) &= \frac{1}{|I_2|} \sum_{i \in I_2} \delta \left( x - \operatorname{mod} (\W^{\prime} \X_i, 2\pi) \right), \label{X2}
\end{align}
where $\operatorname{mod(\cdot, \cdot)}$ is the modular arithmetic, and $\delta(\cdot)$ is the Dirac delta function.
Distributions $p_1(x; \W)$ and $p_2(x; \W)$ are zero when $x \in (-\infty, 0) \cup [2\pi, +\infty)$.
Such distributions are called \emph{circular distributions} or \emph{polar distributions} \citep{Fisher93}.

\subsection{Circular Discrepancy}
We now define a metric for measuring  the discrepancy between $X_1(\W)$ and $X_2(\W)$. Later we will show that this definition is closely related to MMD.

\noindent {\bf Definition~2} \ \ \emph{Given two independent circular distributions $X_1 \sim P_1$ and $X_2 \sim P_2$, we define the \emph{circular discrepancy} as:
\begin{align}
\eta(X_1, X_2) = \sup_{Q} \big( \E_{Q, P_1} \sin(Y - X_1) - \E_{Q, P_2} \sin(Y - X_2) \big), \label{equ:op3}
\end{align}
where $ Y \sim Q$ is also a circular distribution.
}

In this definition, we choose sine function as the measure for assessing the distance between two circular distributions.

\noindent {\bf Claim~5} \ \ \emph{The circular
discrepancy as defined in \eqref{equ:op3} is equal to:
\begin{align}
\eta(X_1, X_2) = \sup_{y \in [0, 2\pi)} \int_0^{2\pi}\! (p_1(x)-p_2(x))\sin(y-x) \, \mathrm{d} x. \label{equ:distr1}
\end{align}
If $p_1(x)$ and $p_2(x)$ are probability mass functions (linear combination of Dirac delta functions), let $p_1(x)-p_2(x) = \sum_{i=1}^N a_i \delta(x-x_i)$, and then the circular discrepancy is equal to
\begin{align}
\eta(X_1, X_2) = \left[\sum_{i=1}^N \sum_{j=1}^N a_i a_j \cos(x_i-x_j) \right]^{\frac{1}{2}}. \label{equ:distr2}
\end{align}
}

The proof is provided in Appendix~A.5.
We can see that the circular discrepancy has a close connection with MMD, see Claim~2.
In fact, if $p_1(x)$ and $p_2(x)$ are probability mass functions, the object function $\int\! (p_1(x)-p_2(x))\sin(y-x) \, \mathrm{d} x$ in Eqn.~\eqref{equ:distr1} is a linear combination of sinusoids with the same frequency. The maximum of this problem is the amplitude of the combined sinusoid, and this is consistent with MMD.

From Eqn.~\eqref{equ:distr1} in Claim~5, it can be seen that the optimal distribution of random variable $Q$ in the definition of circular discrepancy is a Dirac delta function. 
For the integral in Eqn.~\eqref{equ:distr1}, if we change $y$ to $\operatorname{mod} (y+\pi, 2\pi)$, the value of this expression will change its sign. Based on this observation, it is clear that there are two distributions of $Q$ that can maximize and minimize the object function, respectively. The difference of their non-zero positions is $\pi$.
These two distributions construct a \emph{``optimal decision diameter"} for the projected samples on a unit circle.

\begin{figure}[!htbp]
    \centering
    \includegraphics[width=0.6\linewidth]{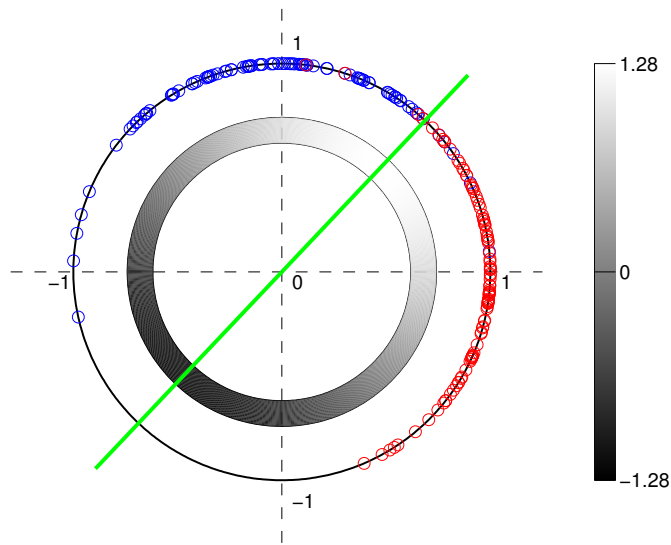}
\caption{Random projection on a unit circle and the decision diameter that maximizes the mean margin. Red and blue circles represent projected samples from two sample sets, respectively. The green line is the decision diameter. The belt represents the value of object function in problem~\eqref{equ:op3} with different angles in $Q$, where light color represents higher value and dark color represents lower value.}
\label{fig:diameter}
\end{figure}

We then give a geometric explanation for problem~\eqref{equ:op3}.
Note that the sine function is a distance measure with sign for two points on a circle
(suppose positive angles are counterclockwise), so the definition of circular discrepancy aims to find a diameter that possibly largely separates the projected samples of two different classes. An example is shown in Fig.~\ref{fig:diameter}, where the orientation of the diameter corresponds to the non-zero elements of distribution $Q$ in the definition of circular discrepancy. We can see that this diameter maximizes the mean margin of two sample classes.

To have a large ensemble of circular discrepancy, the spread parameter $\sigma$ in Gaussian kernel should be neither too small nor two large. If $\sigma \rightarrow 0$, its Fourier transform $p(\W)$ tend to be the uniform distribution. Intuition suggests that if a continuous distribution is spread out over a large region of the line then the corresponding wrapped distribution will be almost uniform on the circle. If $\sigma \rightarrow +\infty$, its Fourier transform $p(\W)$ tend to be Dirac delta function, and all the projected points on the circle would be near zero. In both of the two cases, the ensemble of circular discrepancy is small.
This observation is consistent with the asymptotic behavior of Gaussian kernel SVM \citep{Keerthi03}.

\subsection{Ensemble of Circular Discrepancy}
We have discussed the circular discrepancy for a given random projection $\W$ on a unit circle.
For shift-invariant kernels, $\W$ is not a fixed value, but randomly sampled from a distribution, see Eqn.~\eqref{equ:fmap}.
We use sampling method for ensemble of circular discrepancy under different random projections.
This ensemble turns out to be an efficient approximation of empirical estimate of MMD,
so Algorithm~\ref{alg:alg1} can be explained as ensemble of circular discrepancy.

\noindent {\bf Definition~3} \ \ \emph{Suppose $p(\W)$ is the normalized distribution in Eqn.~\eqref{equ:fmap} for kernel $K$; $X_1$ and $X_2$ are two circular distributions depending on $\W$ according to Eqn.~\eqref{X1}\eqref{X2}. The \emph{ensemble of circular discrepancy} for $X_1$ and $X_2$ with shift-invariant kernel $K$ is defined as:
\begin{align}
 \overline{\eta} (X_1(\W), X_2(\W); p(\W)) = \left[ \E_{\W\sim p(\W)} \ \eta^2(X_1(\W), X_2(\W)) \right]^{\frac{1}{2}}. \label{equ:M3}
\end{align}
}

\noindent {\bf Claim~6} \ \ \emph{
For a shift-invariant kernel  $K(\X,\Y)=\left\langle\phi(\X),\phi(\Y)\right\rangle$, $K(\mathbf{0})=1$; denote $\mathcal{H}$ as the associated Hilbert space with kernel $K$; $p(\W)$ as the normalized distribution in Eqn.~\eqref{equ:fmap}; $X_1$ and $X_2$ as two circular distributions depending on $\W$ defined by Eqn.~\eqref{X1}\eqref{X2}. Then the
ensemble of circular discrepancy is
\begin{align}
\overline{\eta} (X_1(\W), X_2(\W); p(\W)) = \left\| \frac{1}{|I_1|}\sum_{i \in I_1} \phi(\X_i) - \frac{1}{|I_2|}\sum_{i \in I_2} \phi(\X_i) \right\|_{\mathcal{H}}.
\label{equ:theorem2}
\end{align}
}

\noindent{\bf Proof:} \ \ By substituting  Eqn.~\eqref{equ:distr2} into Eqn.~\eqref{equ:M3} and utilizing Bochner's theorem, we can obtain
\begin{align}
\text{(Left hand)}^2
=  & \sum_{i=1}^N \sum_{j=1}^N a_i a_j \E_{\W\sim p(\W)} \cos(\Wt(\X_i-\X_j)) \nonumber \\
=& \sum_{i=1}^N \sum_{j=1}^N a_i a_j K(\X_i, \X_j) = \text{(Right hand)}^2. \nonumber
\label{equ:equiv}
\end{align}
where $a_i = \frac{1}{|I_1|}$ if $i \in I_1$, and $a_i = \frac{-1}{|I_2|}$ if $i \in I_2$.
\qed

\begin{figure}[!htbp]
\begin{center}
{\includegraphics[width=0.8\linewidth]{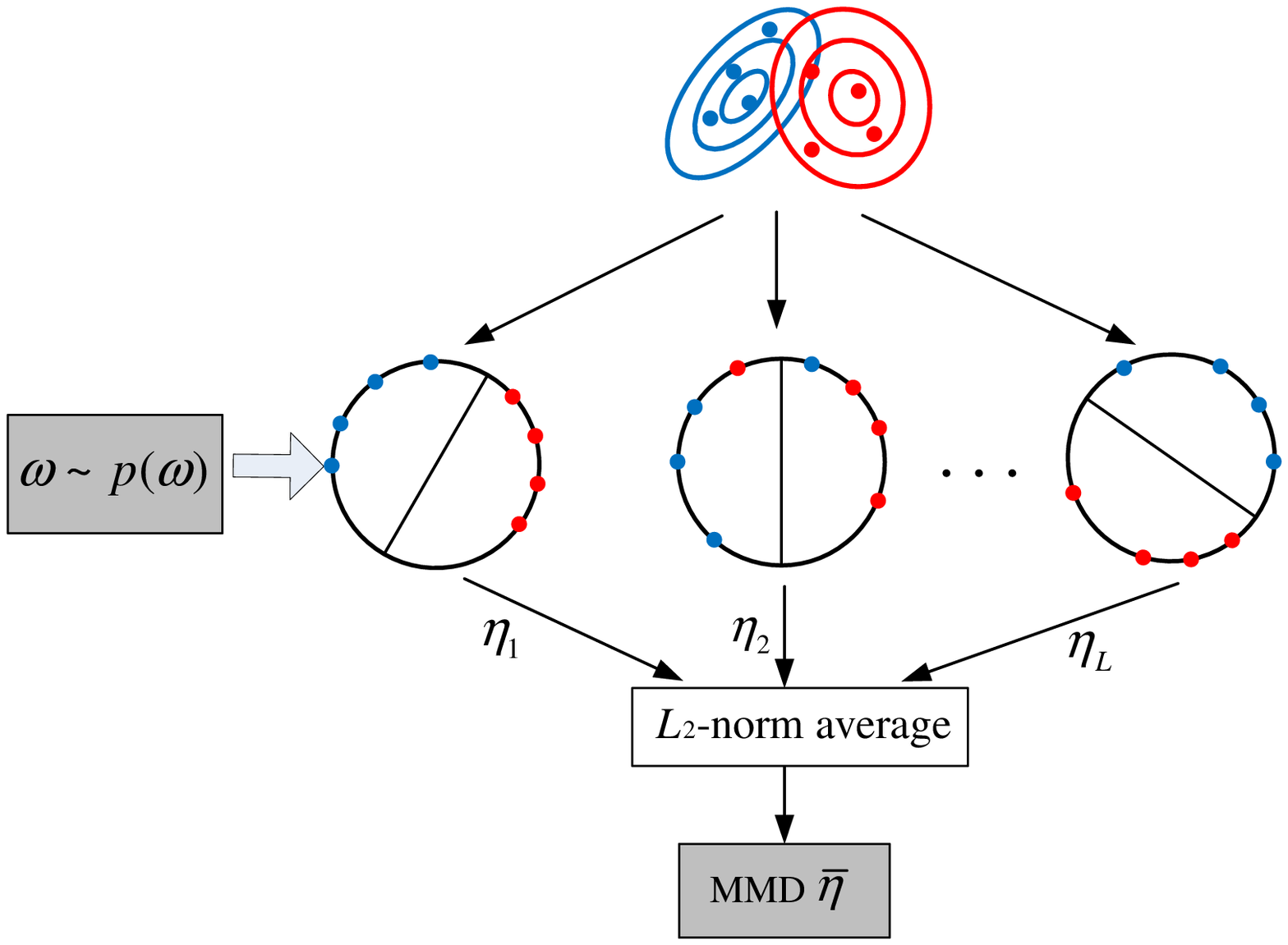}}
\end{center}
\vspace{-0.08in}
\caption{Flowchart of circular discrepancy ensemble.}
\label{fig:flowchart}
\end{figure}

Fig.~\ref{fig:flowchart} demonstrates the relationship between MMD, circular discrepancy and our approximation. The blue and red contours are two distributions, and $p(\W)$ is the distribution determined by Fourier transform of the kernel. We generate i.i.d. samples $\W$ from distribution $p(w)$. For each generated $p(\W)$, we project samples on a unit circle and calculate the circular discrepancy. The ensemble of discrepancy then corresponds to the MMD.
We can see that the circular discrepancy constructs classifiers implicitly.

It is interesting that if we use other similarity measurements such as Mallows Distance (earth mover's distance) \citep{Werman86}, other than the sine function utilized in Definition~2, more extensive metrics for two-sample test can be naturally obtained.
Furthermore, it should be noted that in Definition~3, we use $L_2$-norm average for ensemble. If we use other norms, we can also generalize more measures other than MMD for this task. All these extensions are hopeful directions for future investigation.

\section{Experimental Results}
\subsection{Approximation Quality}
\label{sec:exp1}

\begin{figure}[!htbp]
\begin{center}
    \subfigure[]
    {\includegraphics[width=0.35\linewidth]{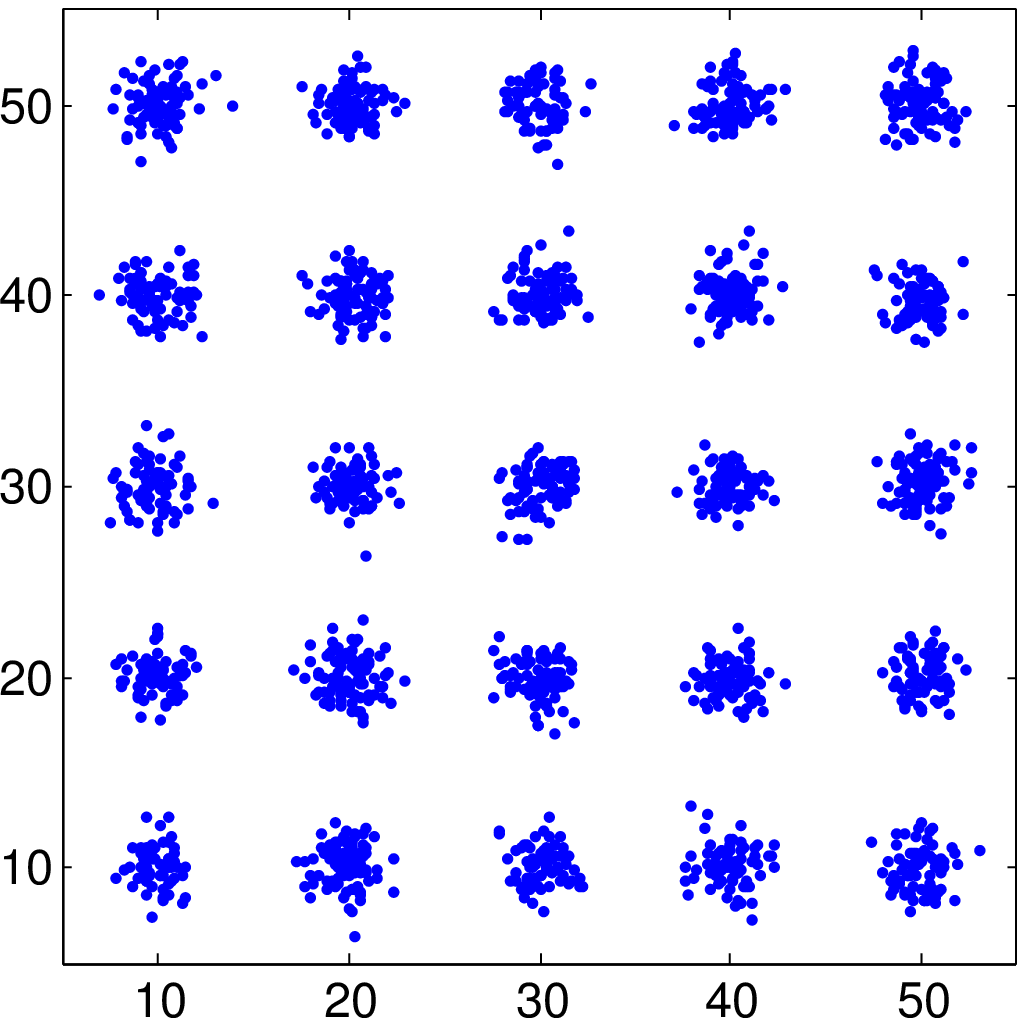}} \quad
    \subfigure[]
    {\includegraphics[width=0.35\linewidth]{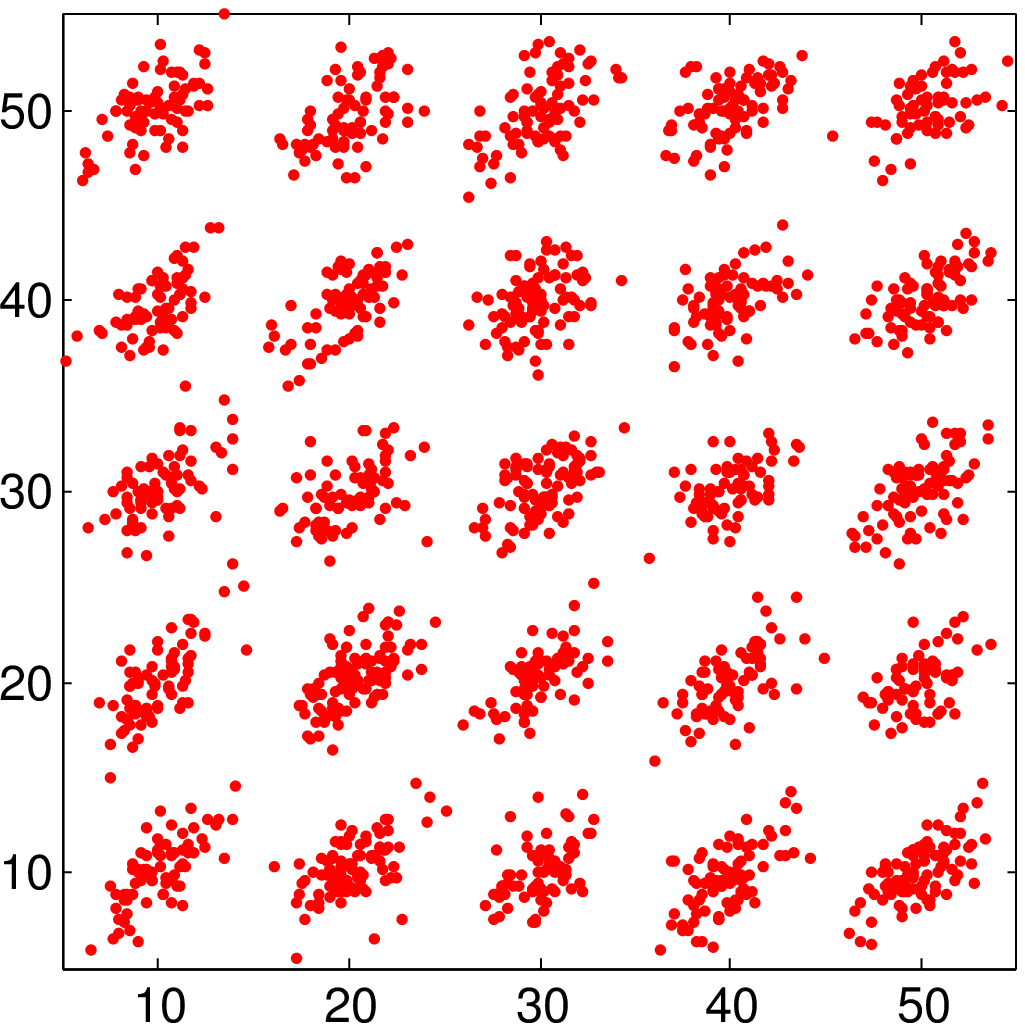}}
\end{center}
\vspace{-0.2in}
\caption{Synthetic data. (a) Distribution $P$; (b) Distribution $Q$.}
\label{fig:syth}
\end{figure}

We begin by investigating how well our methods can approximate the exact MMD  as the sampling number $L$ increases.
Following previous work on kernel hypothesis testing \citep{Gretton12b, Zaremba13}, our synthetic distribution is designed as $5 \times 5$ grids of 2D Gaussian blobs. We specify two distributions, $P$ and $Q$. For distribution $P$ each Gaussian has identity covariance matrix, while for distribution $Q$ the covariance is non-spherical with a ratio $\epsilon$ of large to small covariance eigenvalues. Samples drawn from $P$ and $Q$ are presented in Fig.~\ref{fig:syth}.
The ratio $\epsilon$ is set as $4$, and the sample number for each distribution is set as $1000$.

\begin{figure}[!htbp]
\begin{center}
    \subfigure[biased MMD]
    {\includegraphics[width=0.49\linewidth]{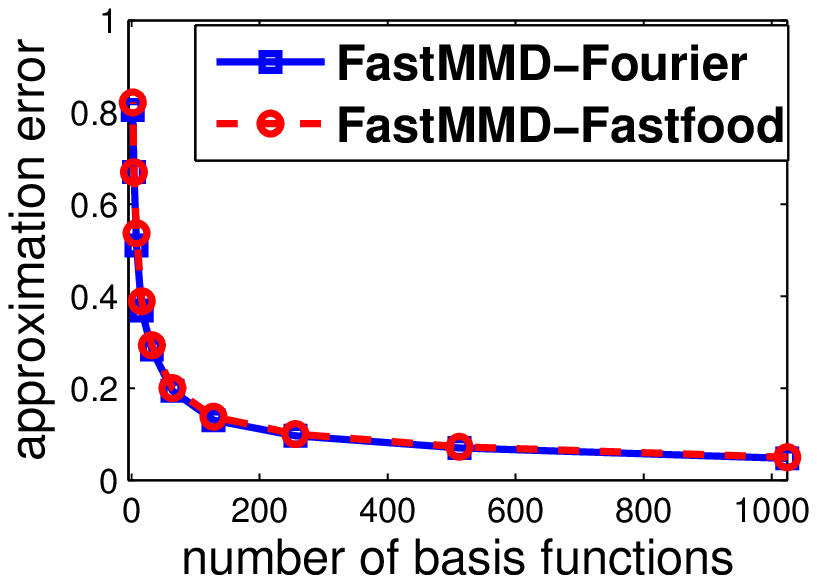}}
    \subfigure[unbiased MMD]
    {\includegraphics[width=0.49\linewidth]{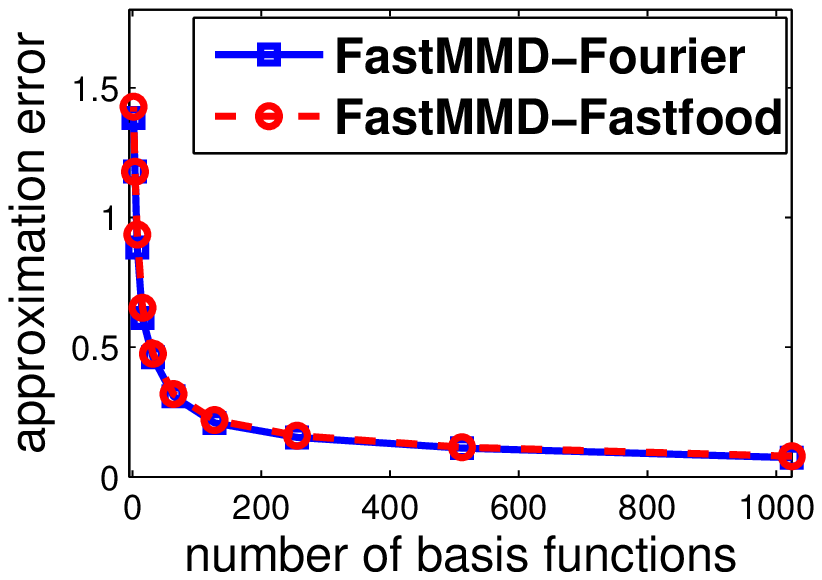}}
\end{center}
\vspace{-0.2in}
\caption{MMD approximation errors of FastMMD with respect to number of basis function $L$.}
\label{fig:appro}
\end{figure}

We used a Gaussian kernel with $\sigma = 1$, which approximately matches the scale of the variance of each Gaussian in mixture $P$.
The MMD approximation results are shown in Fig.~\ref{fig:appro}. We use the relative difference between exact (biased and unbiased) MMD and the approximation to quantify the error. The absolute difference also exhibits similar behavior and is thus not shown due to space limit. The results are presented as averages from $1000$ trials. As can be seen, as $L$ increases, both \emph{FastMMD-Fourier} and \emph{FastMMD-Fastfood} converge quickly to exact MMD in both biased and unbiased cases. Their performance are indistinguishable.
It can be observed that for both methods, the good approximation can be obtained even from a modest number of basis.

\subsection{Accuracy Test for MMD with Kernel Family}
In some situations, MMD with a kernel family is preferred \citep{Sriperumbudur09}. Here we present an experiment that illuminates the superiority of our FastMMD methods on accuracy of MMD with kernel family.
The synthesis data are generated as follows.
All samples are constrained into a two-dimensional rectangle: $-5 \le x_1, x_2 \le 5$. The points, which are located within a circular ring in between by $x_1^2 + x_2^2 = 1$ and $x_1^2 + x_2^2 = 16$ are labeled as $+1$, while other points are labeled as $-1$. We generate $200$ samples for each distribution randomly as the test set. Then we use these samples to calculate MMD with a kernel family. Here the kernel family is composed of multivariate isotropic Gaussians with bandwidth varying between $0.1$ and $100$, with a multiplicative step-size of $10^{1/5}$.

\begin{figure}[!htbp]
\begin{center}
    \subfigure[]
    {\includegraphics[width=0.49\linewidth]{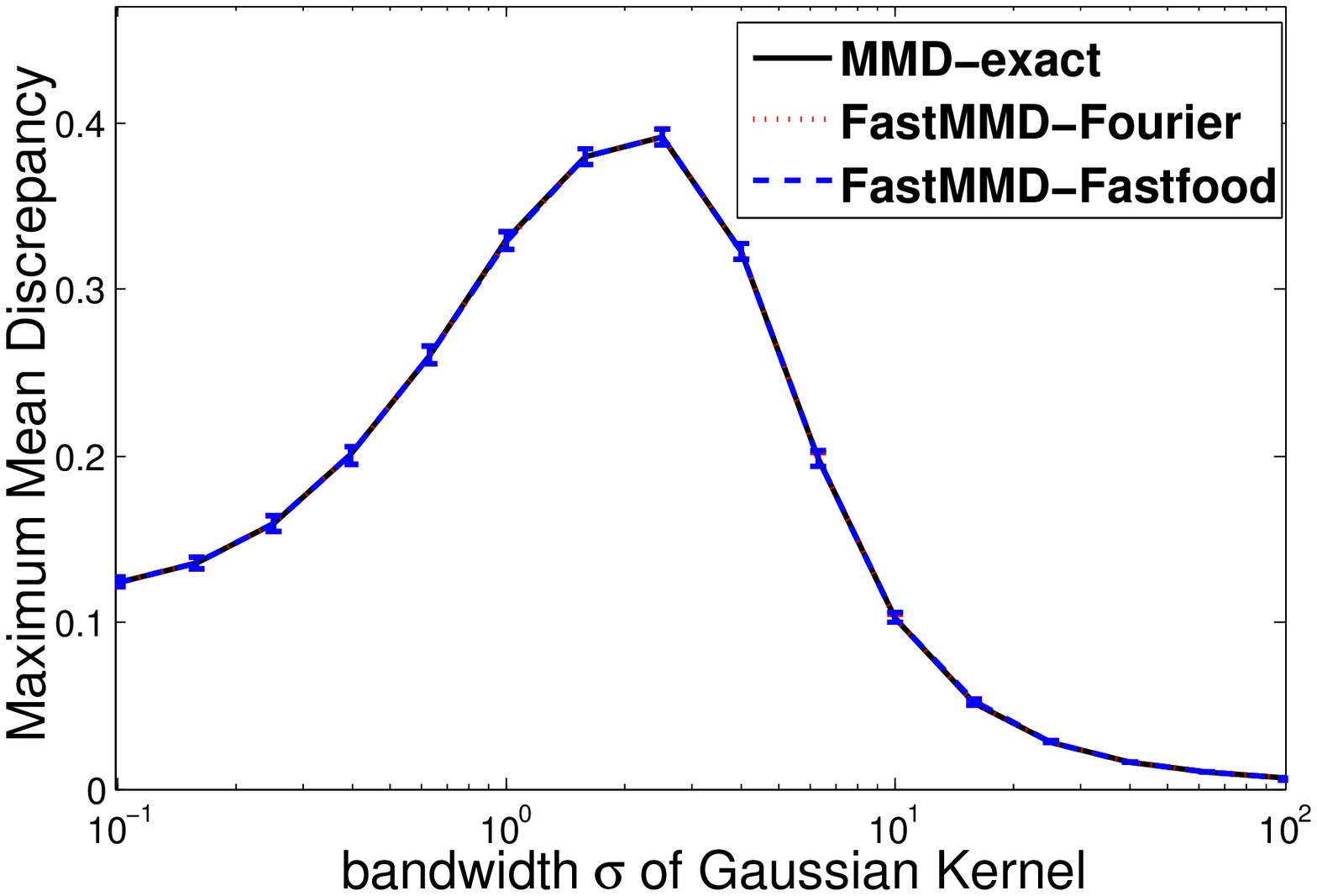}}
    \subfigure[]
    {\includegraphics[width=0.49\linewidth]{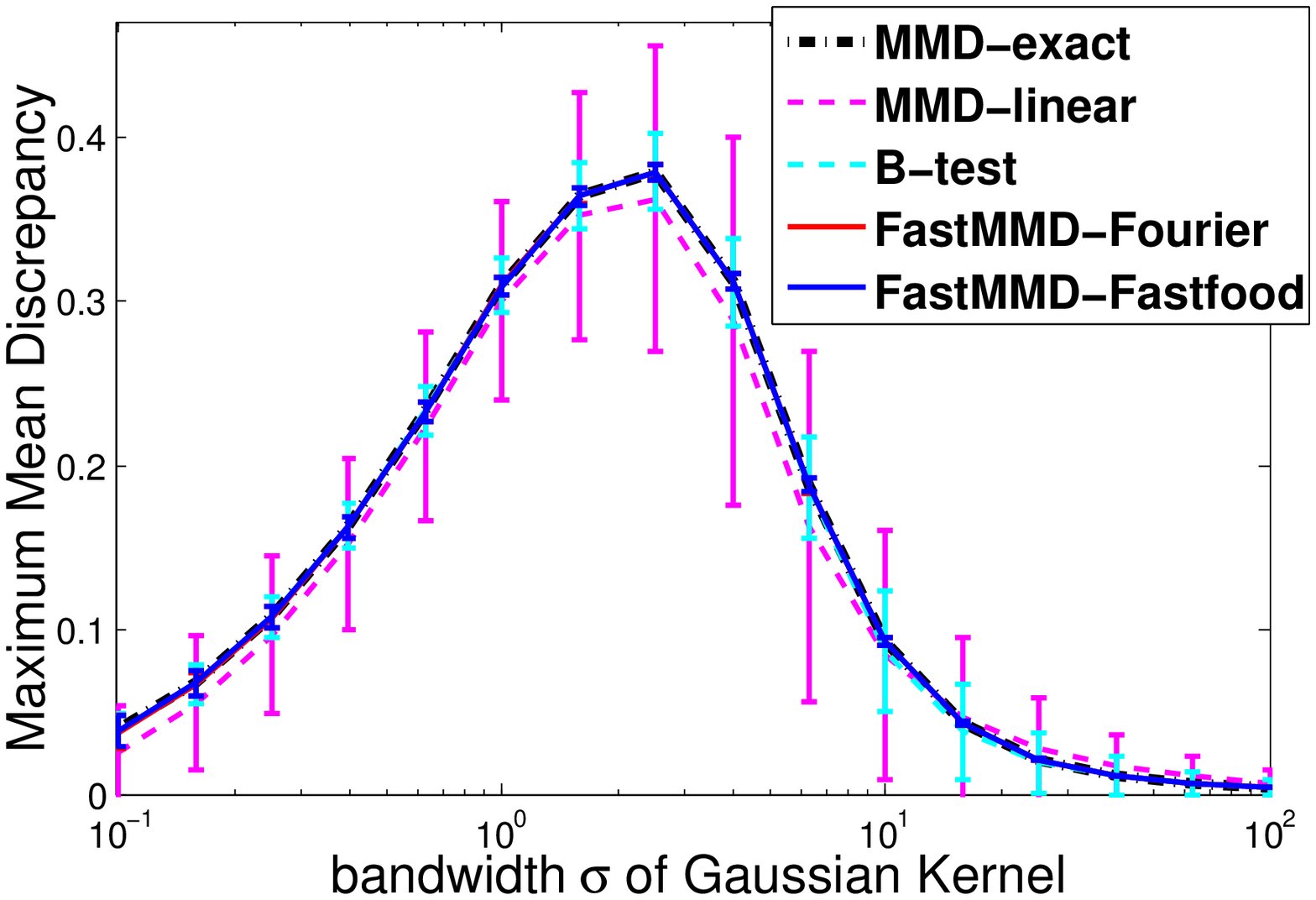}}
    \subfigure[]
    {\includegraphics[width=0.9\linewidth]{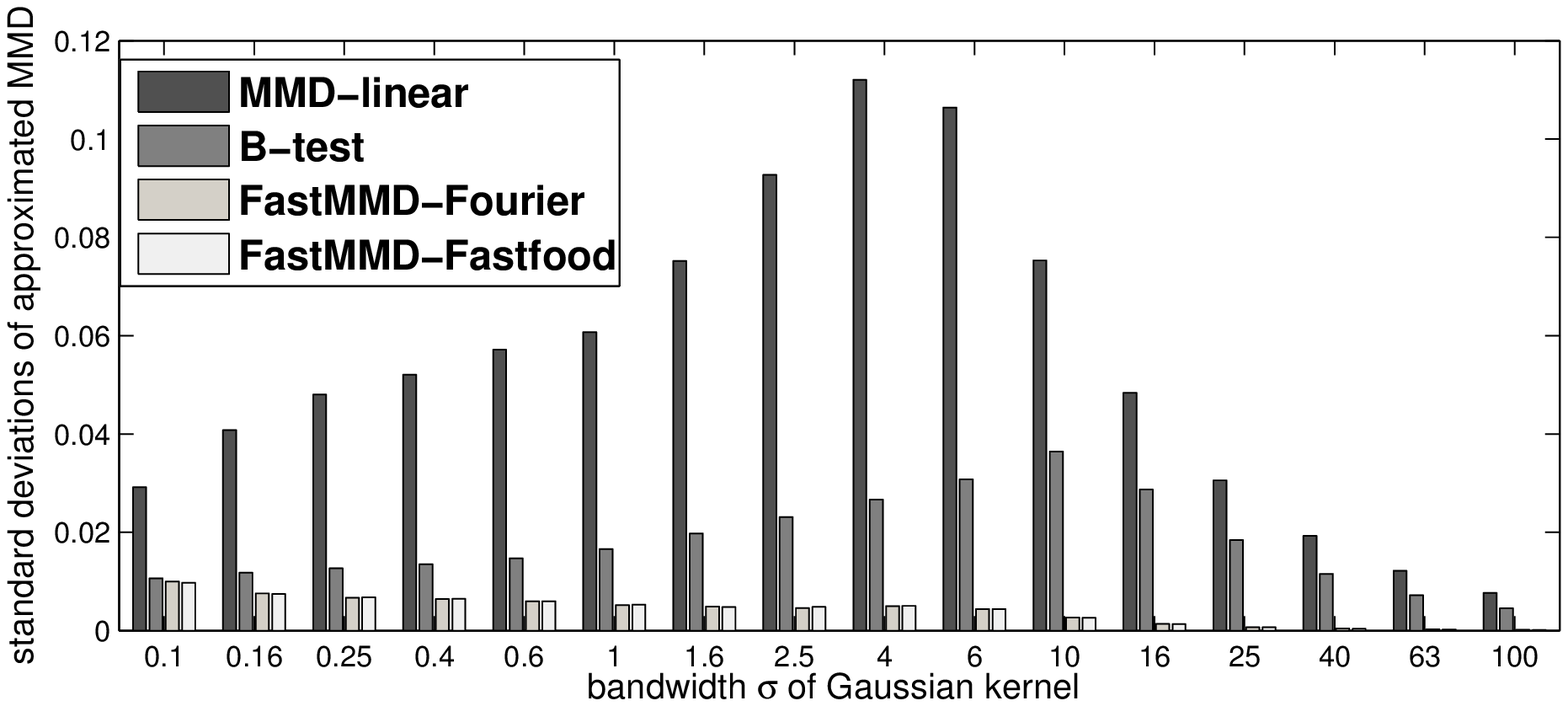}}
\end{center}
\vspace{-0.2in}
\caption{MMD approximation for kernel family with \mbox{different} bandwidth $\sigma$. (a) biased MMD; (b) unbiased MMD; (c) standard deviations of approximated unbiased MMD in (b).}
\label{fig:family}
\end{figure}

We compare our method with exact MMD, \emph{MMD-linear} \citep{Gretton12}, and \emph{B}-test \citep{Zaremba13}. Note that the latter two methods are only valid for unbiased MMD.
In our methods, the number of basis function $L$ is set as $1024$.
The block size in \emph{B}-test is set to the default choice, i.e, the square of sample size $\sqrt{N}$. For our method, we repeat $1000$ times and use the curves and error bars to represent means and standard deviations of MMD, respectively. Since both \emph{MMD-linear} and \emph{B}-test depend on the permutation of data samples, we make $1000$ permutations of the samples. From  Fig.~\ref{fig:family}, it can be seen that the means of all these methods are consistent with the true values. Also it can be seen that
our \emph{FastMMD-Fourier} and \emph{FastMMD-Fastfood} have similar accuracy, and their deviations are much smaller than those of \emph{MMD-linear} and \emph{B}-test.

For some applications, we need to find the kernel that has the maximal MMD. Since our methods have lower variance, they incline to find the correct $\sigma$ with higher probability than \emph{MMD-linear} and \emph{B}-test.

\subsection{Efficiency Test on Synthetic Data}
\begin{figure}[!htbp]
\begin{center}
    \subfigure[]
    {\includegraphics[width=0.49\linewidth]{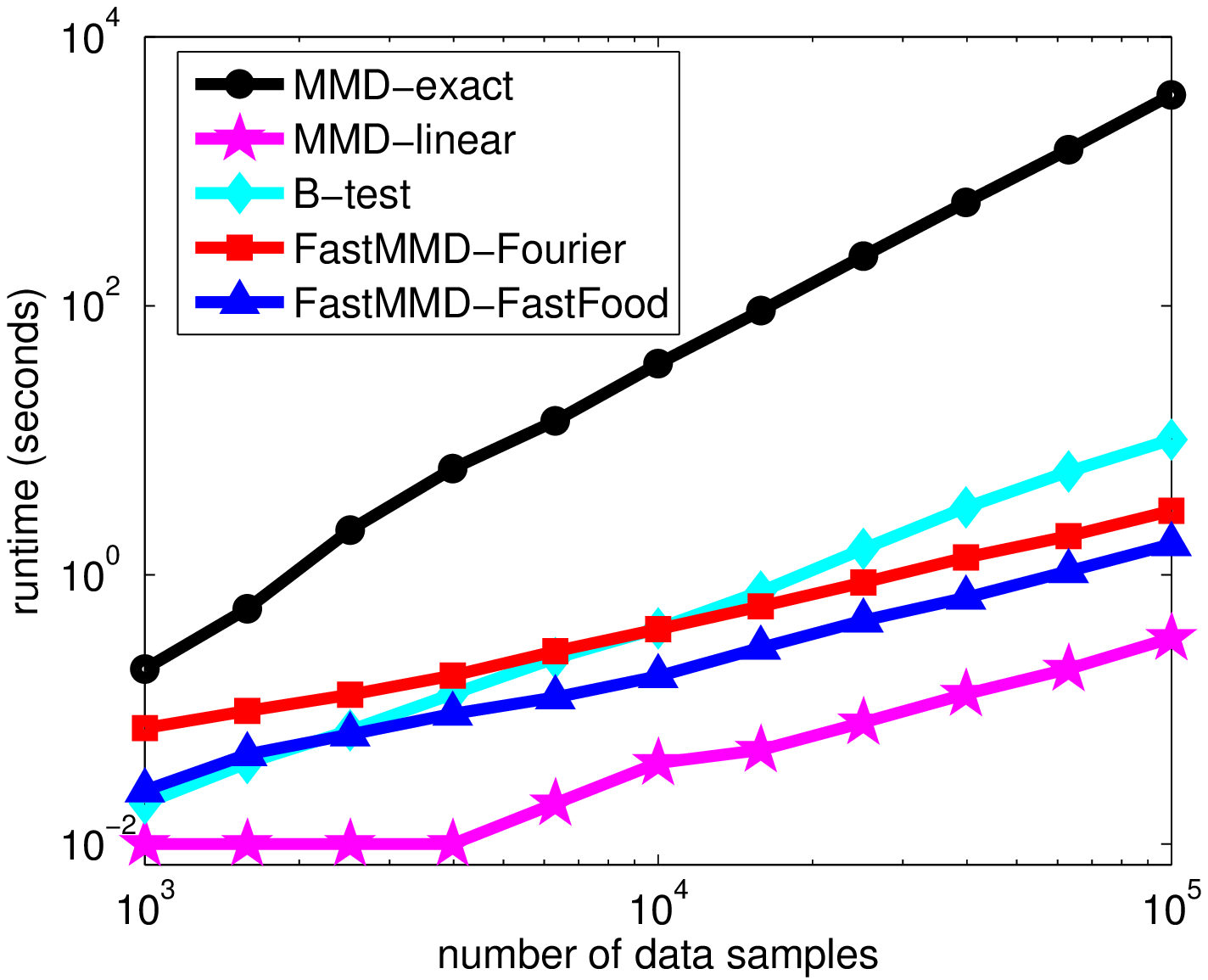}}
    \subfigure[]
    {\includegraphics[width=0.49\linewidth]{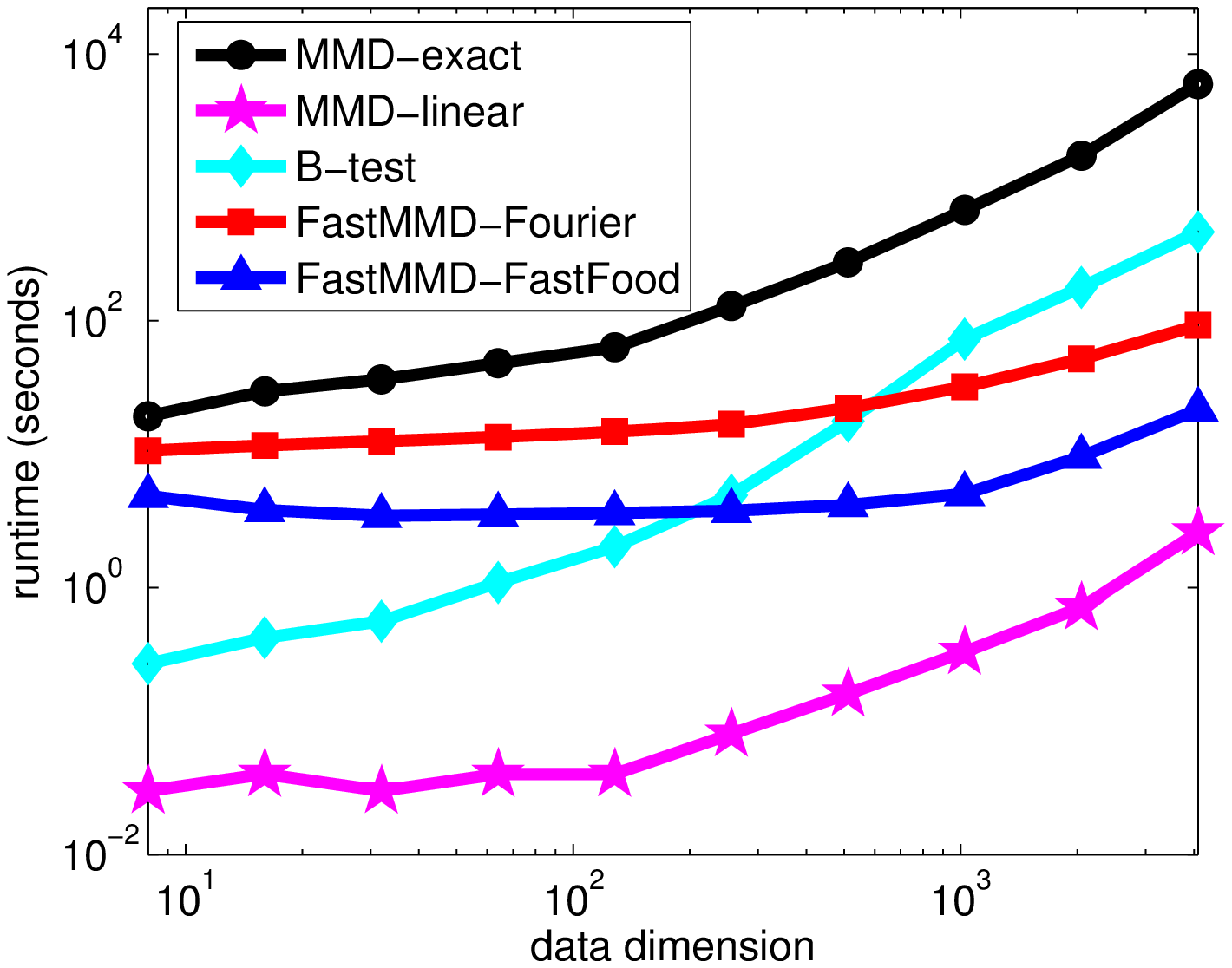}}
\end{center}
\vspace{-0.2in}
\caption{Efficiency comparison of different methods. Our methods have better scalability than the exact solution. (a) Fix $d = 16$ and $L=128$, change number of samples $N$ ; (b) Fix $N = 10^4$ and $L=8192$,  change $d$. Note the axises are log-scale.}
\label{fig:time}
\end{figure}

In order to evaluate the efficiency of our methods, we generate samples uniformly from $[0,0.95]^d$ and $[0.95,1]^d$.
The efficiency of different methods is shown in Fig.~\ref{fig:time}.
In Fig.~\ref{fig:time}(a), the number of samples is varied from $10^3$ to $10^5$, and the data dimension $d$ is set as $16$. In Fig.~\ref{fig:time}(b), the number of samples is set as $10^4$, and  the dimension $d$ is varied from $8$ to $1024$.

All competing methods are implemented in Matlab except that we use Spiral WHT Package\footnote{\url{http://www.spiral.net/}} to perform the Fast Walsh-Hadamard transform in Fastfood.
The comparison methods include exact MMD, \emph{MMD-linear} and \emph{B}-test\footnote{\url{https://github.com/wojzaremba/btest}}.
We run all the codes on a PC with AMD Athlon X2 250 ($800$ MHz) CPU and $4$ GB RAM memory.

From Fig.~\ref{fig:time}, we can see that when $N$ or $d$ varies from small to large, our  methods gradually become more efficient than exact MMD and \emph{B}-test.
When $d =16$ and $N = 10^5$, \emph{FastMMD-Fourier} and \emph{FastMMD-Fastfood} are $2000 / 5$x and $4000 / 10$x faster than exact MMD / \emph{B}-test, respectively.
As for \emph{MMD-linear}, since it is the extreme simplified subsampling version of MMD calculation, it always runs very fast. However, as the sample size or dimension increases, the computation times of our methods still depict a more slowly increase trend than that of \emph{MMD-linear}.
This empirically confirms the efficiency of the proposed FaseMMD methods.

\subsection{Efficiency Test on Large-Scale Real Data}
In order to validate the efficiency of FastMMD on real world data, we also perform experiments on PASCAL VOC 2007 \footnote{\url{http://pascallin.ecs.soton.ac.uk/challenges/VOC/voc2007/}}, which is popular in computer vision community. The dataset consists of $9963$ images, and there are 20 object categories in this dataset with some images containing multiple objects. In our experiments, we choose $4015$ images which contain \textit{persons} as one sample set, and the remaining $5948$ images as another sample set.
We use the \textsc{VLFeat} toolbox\footnote{\url{http://www.vlfeat.org/}} to extract features. For each image, the feature is constructed by bag-of-words representation and spatial pyramid. The codebook size for bag-of-words is $1024$, and the spatial pyramid contains $1\times1$, $2\times2$ and $4\times4$ levels. Thus the feature dimension for each image is $1024 \times (1+2\times2+4\times4) = 21504$. Finally, the feature vectors are normalized by $L_1$ norm.

We use MMD to measure the discrepancy between these two image set. In this MMD calculation, the sample number is $N=9963$, and data dimension is $d=21504$. The number of basis function $L$ is fixed as $1024$.
The bandwidth $\sigma$ is set as $10^{-2.2} = 0.0631$.
We perform this experiment on a PC with Intel Core i7-3770 $@3.4$ GHz CPU and $32$ GB RAM memory.

The results of efficiency and accuracy are shown in Tab.~\ref{table:pascal}.
We compare our method with exact MMD, \emph{MMD-linear} and \emph{B}-test. Except for exact MMD, we repeat $10$ times for each method, and report the mean and standard deviation of MMD, execution time, and relative error. We can see that our FastMMD methods have smaller approximation error and smaller deviation, and meanwhile it has three orders of speedup compared to exact MMD.

\setlength{\tabcolsep}{4pt}
\begin{table}[!htbp]
\begin{center}
\caption{Efficiency and accuracy comparison on PASCAL VOC 2007.}
\label{table:pascal}
\begin{tabular*}{0.99\textwidth}{@{\extracolsep{\fill}}lccccc}
\toprule
 & Exact & MMD-linear & B-test & MMD-Fourier & MMD-Fastfood \\
\midrule
MMD mean & 0.1084 & 0.0817 & 0.1081 & 0.1083 &  0.1083 \\
std. dev. & - & 5.276$\times 10^{-2}$ & 5.168$\times 10^{-3}$ & $\text{\bf 1.150} \mathbf{\times} \text{\bf 10}^{\text{\bf -3}}$ & 1.261$\times 10^{-3}$ \\
relative err. & - & 24.6\%  & 0.277\% & {\bf 0.0923\%} & {\bf 0.0923\%} \\
\midrule
time / second & 15590.1 & 185.5 & 211.7 & 13.08 & {\bf 6.45} \\
speedup & - & 84  & 74  & 1192  & {\bf 2417} \\
\bottomrule
\end{tabular*}
\end{center}
\end{table}
\setlength{\tabcolsep}{1.4pt}

\subsection{Type II Error}
Given MMD, several strategies can be employed to calculate the test threshold \citep{Gretton09, Gretton12, Chwialkowski14}.
The bootstrap strategy \citep{Arcones92} is utilized in our experiments since it can be easily integrated into our FastMMD method.
Also the bootstrap is preferred for large-scale datasets since it costs $O(N^2 d)$, faster than most other methods for this task such as Person, with cost $O(N^3 d)$ \citep{Gretton12}.

The data used for this experiment is generated from Gaussian blob distributions as described in Section~\ref{sec:exp1}.
The sample size is set as $1000$ for two distributions.
The bandwidth is selected by maximizing MMD. The selected bandwidth is $\sigma =1$, and it approximately matches the scale of variance of each Gaussian blob in distribution $P$. We find that all of biased/unbiased FastMMD methods have similar good performance, and we thus only demonstrate the result of \emph{FastMMD-Fourier} for biased MMD.

\begin{figure}[!htbp]
\begin{center}
    \subfigure[]
    {\includegraphics[width=0.485\linewidth]{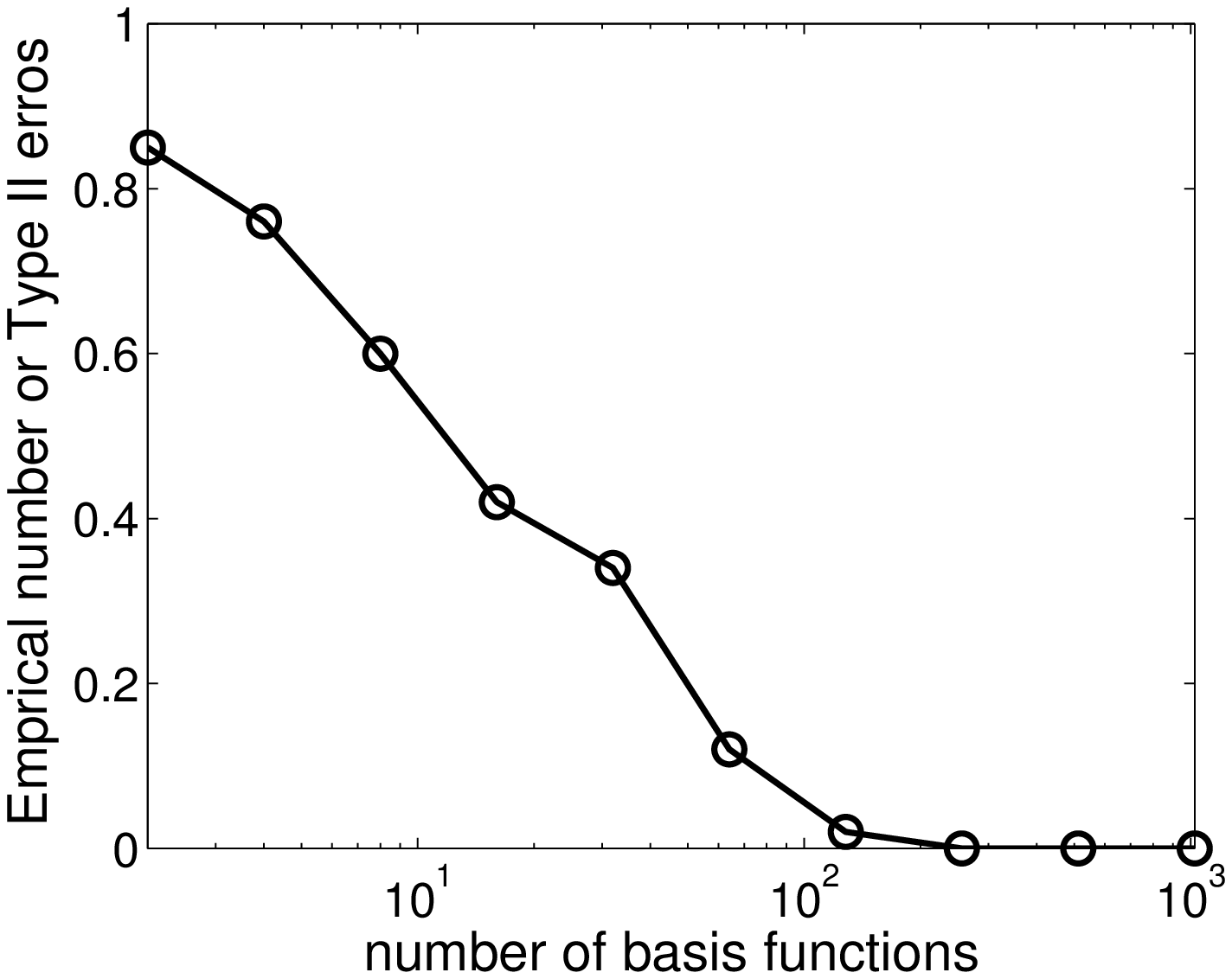}}
    \subfigure[]
    {\includegraphics[width=0.49\linewidth]{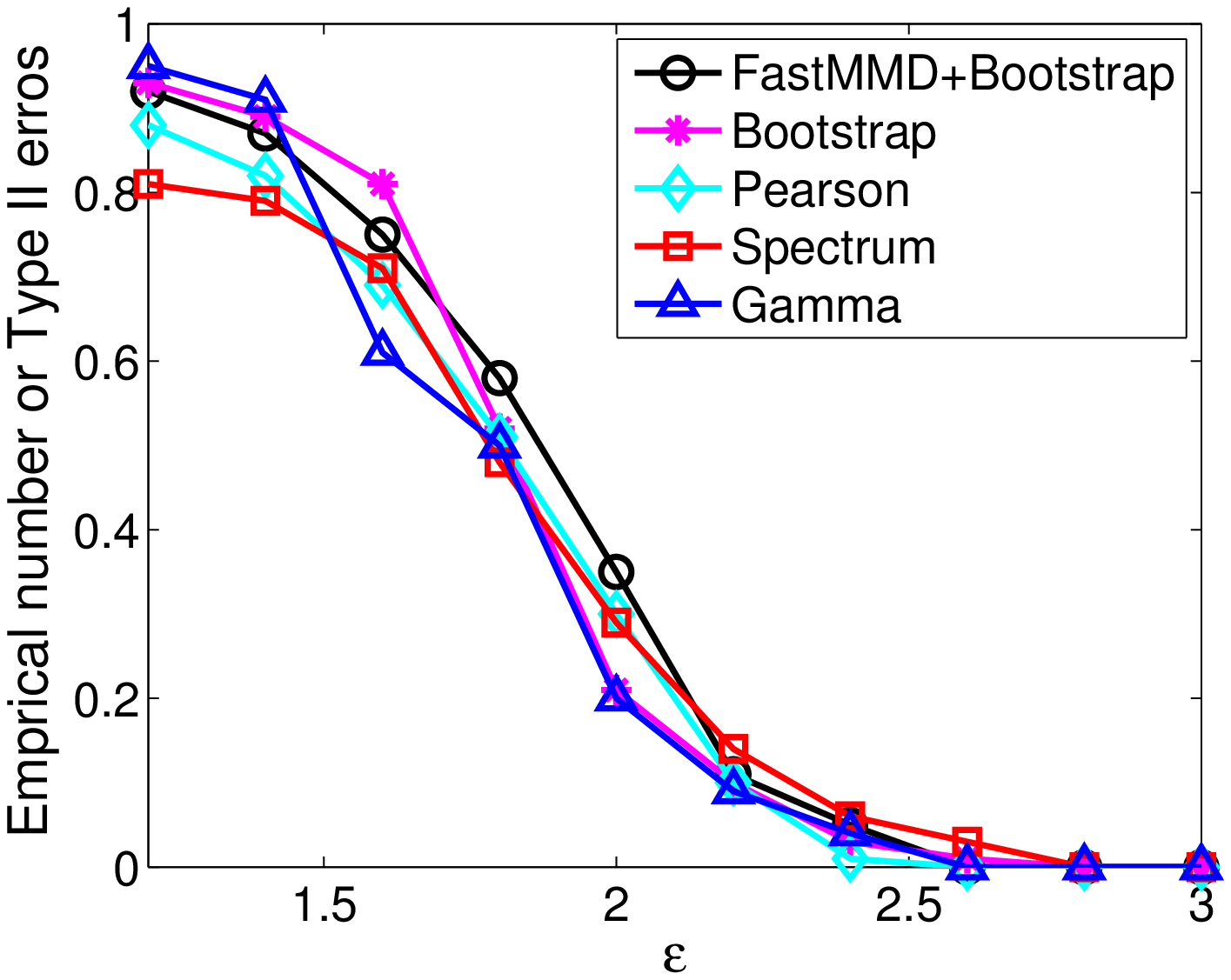}}
\end{center}
\vspace{-0.2in}
\caption{Type II error when using \emph{FastMMD-Fourier} for biased MMD. (a) Fix $\epsilon = 4$, change $L$; (b) Fix $L = 256$ and the ratio $\epsilon$ of variance eigenvalues is varied from $1.2$ to $3.0$. Results from other four typical methods are also included for comparison, and each result is the average of $100$ trials.}
\label{fig:typeII}
\end{figure}

The level $\alpha$ for Type I error is set as $0.05$, and the number of bootstrap shuffles is $1000$. The Type II error is shown in Fig.~\ref{fig:typeII}(a). We can see that the Type II error drops quickly when increasing number of basis. It demonstrates empirically that increasing number of basis can decrease the Type II error.

We also compare our method with \textit{Bootstrap} (Bootstrap approach with exact MMD), \textit{Pearson} (moment matching to Pearson curves), \textit{spectrum} (Gram matrix eigenspectrum) and \textit{Gamma} (two-parameter Gamma approximation), where the former two approaches are presented in \citet{Gretton12} and the latter two are introduced in \cite{Gretton12b}. The basis number utilized in FastMMD is fixed as $256$.
The type II errors of all competing methods with respect to $\epsilon$ are shown in Fig.~\ref{fig:typeII}(b).
The execution times for different approaches are $88.7$s (\textit{Bootstrap}), $37.9$s (\textit{Pearson}), $33.2$s (\textit{spectrum}), $0.73$s (\textit{Gamma}), and $4.5$ (our FastMMD+Bootstrap). We can see that the proposed FastMMD method can achieve comparable results with other methods, while it is evidently more efficient than \textit{Bootstrap}, \textit{Pearson} and \textit{spectrum}. This actually substantiates the efficiency of FastMMD since our method and \textit{Bootstrap} are the same except with different strategies for calculating MMD. Our method is less efficient than \textit{Gamma} method because bootstrap methods need to calculate MMD for many times (it is determined by the number of bootstrap shuffles which is $1000$ in this experiment). How to design effective statistic based on FastMMD and optimal threshold remains a challenging issue for future investigation.

\section{Conclusions}
In this paper, we propose a method, called FastMMD, for efficiently calculating the maximum mean discrepancy (MMD) for shift-invariant kernels.
Taking advantage of Fourier transform of shift-invariant kernels, we get a linear time complexity approximation method. We prove the theoretical convergence of the proposed method in both unbiased and biased cases, and further present a geometric explanation for it. This explanation on one side delivers new insight for intrinsic MMD mechanism, and on the other side is hopeful to inspire more extensive new metrics for two-sample test, which will be investigated in our future research.
Future work also includes finding the significance threshold using  more efficient and effective strategies other than bootstrap.

\section*{Acknowledgement}
We thank Alex Smola for valuable suggestions. We also thank anonymous reviewers for their constructive comments, which helped us to improve the manuscript. This research was supported by the National Basic Research Program of China (973 Program) under Grant No. 2013CB329404, the China NSFC project under contract 61373114, 11131006, and the Civil Aviation Administration of China jointly funded project No. U1233110.

\section*{Appendix}
\subsection*{A.1 Relationship between biased and unbiased estimate of MMD}

The unbiased estimate of MMD is (See Lemma 6 in \citet{Gretton12})
\begin{align}
\text{\rm MMD}_{\text{u}}^2[\mathcal{F}, X_1, X_2] = & \frac{1}{m(m-1)} \sum_{i \in I_1} \sum_{j \in I_1, j \ne i} K(\X_i, \X_j) + \nonumber \\
& \frac{1}{n(n-1)} \sum_{i \in I_2} \sum_{j \in I_2, j \ne i} K(\X_i, \X_j) -\frac{2}{mn} \sum_{i \in I_1} \sum_{j \in I_2} K(\X_i, \X_j),\nonumber
\end{align}
where $m = |I_1|$, $n = |I_2|$.

Denote $S_1 = \frac{1}{m^2} \sum_{i \in I_1} \sum_{j \in I_1} K(\X_i, \X_j)$, $S_2 = \frac{1}{n^2} \sum_{i \in I_2} \sum_{j \in I_2} K(\X_i, \X_j)$. If the kernel $K$ is shift-invariant, let $k_0 = K(\mathbf{0}, \mathbf{0}) = K(\X_i, \X_i)$ for any $\X_i$, and then we have
\begin{align*}
\text{MMD}_{\text{u}}^2 &= (\text{MMD}_{\text{b}}^2 - S_1 - S_2) + \frac{m}{m-1} (S_1 - k_0/m) + \frac{n}{n-1} (S_2 - k_0/n) \nonumber \\
&= \text{MMD}_{\text{b}}^2 + \frac{1}{m-1} S_1 + \frac{1}{n-1} S_2 - \frac{m+n-2}{(m-1)(n-1)} k_0.
\end{align*}

Denote $\mu_1 = \frac{1}{m} \sum_{i \in I_1} \phi(\X_i)$ and $\mu_2 = \frac{1}{n} \sum_{i \in I_2} \phi(\X_i)$.
The biased and unbiased estimate of MMD can be reformulated as:
\begin{align}
\text{\rm MMD}_{\text{b}}^2 &= \left\|
\mu_1 - \mu_2
\right\|_{\mathcal{H}}^2, \\
\text{MMD}_{\text{u}}^2 &= \left\| \mu_1 - \mu_2 \right\|_{\mathcal{H}}^2 +  \frac{1}{m-1} \| \mu_1 \|_{\mathcal{H}}^2 + \frac{1}{n-1} \| \mu_2 \|_{\mathcal{H}}^2  - \frac{m+n-2}{(m-1)(n-1)} k_0.
\end{align}

\subsection*{A.2 Preliminary for the Proof of Theorem~3 and Theorem~4}

The following two claims and an equivalent form of Algorithm~\ref{alg:alg1} provide basic tools to analyze the MMD approximation.

\vspace{0.2in}

\noindent {\bf Claim~7 (Uniform Convergence for Linear Combination of Approximation)} \ \ \emph{Let $\M$ be a compact subset of ${\bbbr}^d$, $f(\cdot; \W)$ be a function with parameter $\W$.
If $\W$ is drawn from a distribution, and the difference of two functions is bounded as
\begin{equation}
\text{\rm Pr}\left[\sup_{\X_i, \Y_j \in \M} \left|f(\X_i, \Y_j; \W) - g(\X_i, \Y_j) \right| \ge \epsilon \right] \le C(\epsilon), \label{equ:cond1}
\end{equation}
where $C(\epsilon)$ is function about $\epsilon$. Then
\begin{align}
\text{\rm Pr}\left[\sup_{\{\X_i\}, \{\Y_j\} \subset \M} \left| \sum_{(i,j)} a_i a_j f(\X_i, \Y_j; \W) - \sum_{(i,j)} a_i a_j g(\X_i, \Y_j) \right| \ge \epsilon \right]
\le C\left(\frac{\epsilon}{\sum_{(i,j)} \left| a_i a_j \right|} \right).
\end{align}
}

\noindent{\bf Proof: } For any fixed $\W$, we have
\begin{align*}
& \sup_{\{\X_i\}, \{\Y_j\} \subset \M} \left| \sum_{(i,j)} a_i a_j f(\X_i, \Y_j; \W) - \sum_{(i,j)} a_i a_j g(\X_i, \Y_j) \right| \nonumber \\
\le & \sup_{\{\X_i\}, \{\Y_j\} \subset \M} \sum_{(i,j)} \left| a_i a_j \right| \cdot \left| f(\X_i, \Y_j; \W) - g(\X_i, \Y_j) \right| \nonumber \\
\le & \sum_{(i,j)} \left| a_i a_j \right| \cdot \sup_{\X_i, \Y_j \in \M} \left| f(\X_i, \Y_j; \W) - g(\X_i, \Y_j) \right|.
\end{align*}
Based on Eqn.~\eqref{equ:cond1}, we know that with probability more than $1-C(\epsilon)$, it hold that $\sup_{\X_i, \Y_j \in \M} \left| f(\X_i, \Y_j; \W) - g(\X_i, \Y_j) \right| < \epsilon$. So with probability more than $1-C(\epsilon)$, it holds that $ \sup_{\{\X_i\}, \{\Y_j\} \subset \M} \left| \sum_{(i,j)} a_i a_j f(\X_i, \Y_j; \W) - \sum_{(i,j)} a_i a_j g(\X_i, \Y_j) \right|$ $< \sum_{(i,j)} \left| a_i a_j \right| \cdot \epsilon $. The proof is then completed.
\qed

\noindent {\bf Claim~8} \ \ \emph{
The following inequality holds
\begin{align}
\exp( a - \sqrt{a^2+b^2/x^2} ) < \sqrt{2} a b^{-2} x^2, \quad x>0, a>1, b>0.
\end{align}
}
\noindent{\bf Proof: }
We only need to prove
\begin{align*}
\exp( a - \sqrt{a^2+x^{-2}} ) < \sqrt{2} a x^2, \quad x>0, a>1,
\end{align*}
and then make a variable substitution $ x \leftarrow x^\prime/b$.

Taking logarithm of both sides we have
\begin{align*}
a - \sqrt{a^2+x^{-2}} < \log (\sqrt{2} a) +  2 \log x.
\end{align*}
Define a function $f(x) = 2 \log x + \sqrt{a^2 + x^{-2}} + \log (\sqrt{2}a) - a$. Next we prove three properties of $f(x)$.

(i) $\lim_{x \rightarrow 0^+} f(x) = +\infty$, because $x^{-2}$ dominates the other when $x \rightarrow 0^+$. Also it is obvious that $\lim_{x \rightarrow +\infty} f(x) = +\infty$.

(ii) $f^\prime(x) = \frac{1}{x} \left(2- \frac{1}{x^2 \sqrt{a^2+x^{-2}}} \right)$. Let $f^\prime(x^*)=0$, and then we obtain the only solution $x^* = \sqrt{\frac{-1+\sqrt{1+a^2}}{2a^2}}$.
Since $f(x)$ is continuous and differentiable, $f(0^+)=+\infty$, and $f(+\infty) = +\infty$, we have that $f(x^*)$ is the minimum of this function.

(iii) Finally, we prove $f(x^*)>0$.
\begin{align*}
f(x^*) = \log(-1+\sqrt{1+a^2}) - \log(\sqrt{2} a) + (1+\sqrt{1+a^2}) - a.
\end{align*}
Denote $g(a) = \log(-1+\sqrt{1+a^2}) - \log(\sqrt{2} a) + (1+\sqrt{1+a^2}) - a$. We can verify that $g^\prime(a) = \sqrt{1+a^{-2}} - 1>0$ and $g(1)>0$. So $f(x^*) = g(a)>0$.
\qed

\noindent{\bf An equivalent implementation of FastMMD: }

We provide an equivalent form of FastMMD.
It is easy to verify that
\begin{align}
& \sum_{i=1}^N \sum_{j=1}^N a_i a_j \cos(\Wt\X_i-\Wt\X_j) \nonumber \\
= & \sum_{i=1}^N \sum_{j=1}^N a_i a_j \left[ \cos(\Wt\X_i)\cos(\Wt\X_j) + \sin(\Wt\X_i)\sin(\Wt\X_j) \right] \nonumber \\
= & \left( \sum_{i=1}^N a_i \cos \Wt\X_i \right)^2 + \left( \sum_{i=1}^N a_i \sin \Wt\X_i \right)^2
= \left\| \sum_{i=1}^N \mathbf{v}(\X_i) \right\|_2^2,
\end{align}
where $\mathbf{v}(\X_i) = \left[ \cos(\Wt_1 \X_i), \cdots, \cos(\Wt_L \X_i), \sin(\Wt_1 \X_i), \cdots, \sin(\Wt_L \X_i) \right]$.

The time complexity of this method is also linear. Note that the expression on the right hand is the square length for weighted addition of random Fourier features \citep{Rahimi07}, which means this method uses the random Fourier features to calculate MMD.
The procedure for approximating MMD is described in Algorithm~\ref{alg:alg2}.
Note that this algorithm is equivalent to Algorithm~\ref{alg:alg1} as described in the maintext. Algorithm~\ref{alg:alg1} provides us a geometric explanation of MMD, and Algorithm~\ref{alg:alg2} is convenient for the following uniform convergence analysis.

\begin{algorithm}[!htbp]
   \caption{Approximation of maximum mean discrepancy for shift-invariant kernels} \label{alg:alg2}
   \textbf{Input:} Sample set $S = \{ ( \X_i, \ell_i ) \}^{N}_{i=1}$; shift-invariant kernel $K(\boldsymbol{\Delta})$. \\
   \textbf{Output:} MMD approximation $\overline{\text{\rm MMD}}_{\text{b}}^2$, $\overline{\text{\rm MMD}}_{\text{u}}^2$.
\begin{algorithmic}[1]
   \STATE Calculate the Fourier transform $\mu(\W)$ of $K(\boldsymbol{\Delta})$, and set $p(\W) = \mu(\W)/K(\mathbf{0})$.
   \STATE Draw $L$ samples $\{\W_k\}^{L}_{k=1}$ from $p(\W)$.
   \FOR{$i=1$ {\bfseries to} $N$}
   \STATE $\Z(\X_i) = \frac{K(\mathbf{0})}{\sqrt{L}} \left[ \cos(\Wt_1 \X_i), \cdots, \cos(\Wt_L \X_i), \sin(\Wt_1 \X_i), \cdots, \sin(\Wt_L \X_i) \right] $.
   \ENDFOR
   \STATE $\Z_1 = \frac{1}{|I_1|} \sum_{i \in I_1} \Z(\X_i)$; $\Z_2 = \frac{1}{|I_2|} \sum_{i \in I_2} \Z(\X_i)$.
   \STATE $\overline{\text{\rm MMD}}_{\text{b}}^2 = \| \Z_1 - \Z_2 \|^2$.
   \STATE $\overline{\text{\rm MMD}}_{\text{u}}^2 = \| \Z_1 - \Z_2 \|^2 + \frac{1}{|I_1|-1} \| \Z_1 \|^2 + \frac{1}{|I_2|-1} \| \Z_2 \|^2 - \frac{|I_1|+|I_2|-2}{(|I_1|-1)(|I_2|-1)} K(\mathbf{0})$.
\end{algorithmic}
\end{algorithm}

\subsection*{A.3 Proof of Theorem~3}

From Algorithm~\ref{alg:alg2}, it is easy to see that:
\begin{align}
\overline{\text{\rm MMD}}_{\text{b}}^2 = \| \Z_1 - \Z_2 \|^2 = \sum_{i=1}^N \sum_{i=1}^N a_i a_j \Z'(\X_i; \W) \Z(\X_j; \W),
\label{equ:app1}
\end{align}
where  $a_i = \frac{1}{|I_1|}$ if $i \in I_1$ and $a_i = \frac{1}{|I_2|}$ if $i \in I_2$,  $\W = \{ \W_i \}_{i=1}^L$. $K(\cdot)$ and $\Z(\X_i; \W)$ are the same as that in Algorithm~\ref{alg:alg2}.

According to Eqn.~(5) in \citet{Gretton12}:
\begin{align}
\text{MMD}_b^2 = \sum_{i=1}^N \sum_{i=1}^N a_i a_j K(\X_i, \X_j).
\label{equ:app2}
\end{align}
Based on the above Eqn.~\eqref{equ:app1}\eqref{equ:app2}, we have:
\begin{align}
& \text{\rm Pr}\left[\sup_{\X_1, \cdots \X_N \in \M} \left| \overline{\text{\rm MMD}}_{\text{b}}^2 - \text{\rm MMD}_{\text{b}}^2 \right| \ge \epsilon \right] \nonumber \\
= & \text{\rm Pr}\left[\sup_{\X_1, \cdots \X_N \in \M} \left| \sum_{i=1}^N \sum_{i=1}^N a_i a_j \Z'(\X_i; \W) \Z(\X_j; \W) - \sum_{i=1}^N \sum_{i=1}^N a_i a_j K(\X_i, \X_j) \right| \ge \epsilon \right]
\label{equ:ieq1}
\end{align}

According to Claim~1 in Random Features \citep{Rahimi07}:
\begin{align}
& \text{\rm Pr}\left[\sup_{\X_i, \X_j \in \M} \left| \Z'(\X_i; \W) \Z(\X_j; \W) - K(\X_i, \X_j) \right| \ge \epsilon \right] \nonumber \\
\le & 2^{8} \left( \frac{\sigma_p \operatorname{\text{\rm diam}}(\M)}{\epsilon} \right)^2 \exp\left( -\frac{L \epsilon^2}{4(d+2)} \right)
\triangleq C (\epsilon).
\label{equ:ieq2}
\end{align}

Denote the right hand of the inequality~\eqref{equ:ieq2} as $C(\epsilon)$. Based on Claim~7, the Eqn.~\eqref{equ:ieq1} is bounded by $C\left( \frac{\epsilon}{\sum_{i=1}^N \sum_{j=1}^N |a_i a_j|} \right) = C\left(\frac{\epsilon}{4} \right)$, then we can obtain the uniform convergence for biased estimate of MMD.

Next we prove the case for unbiased estimate of MMD.
According to Algorithm~\ref{alg:alg2} and by virtue of certain algebraically equivalent transformation, we obtain:
\begin{align}
& \overline{\text{\rm MMD}}_{\text{u}}^2 =
\frac{1}{|I_1|(|I_1|-1)} \sum_{i \in I_1} \sum_{j \in I_1, j \ne i} \Z'(\X_i; \W) \Z(\X_j; \W) + \nonumber \\
& \frac{1}{|I_2|(|I_2|-1)} \sum_{i \in I_2} \sum_{j \in I_2, j \ne i} \Z'(\X_i; \W) \Z(\X_j; \W) -\frac{2}{|I_1| \cdot |I_2|} \sum_{i \in I_1} \sum_{j \in I_2} \Z'(\X_i; \W) \Z(\X_j; \W). \label{equ:ieq3}
\end{align}

According to Eqn.~(3) in \citet{Gretton12}, we know that
\begin{align}
\text{\rm MMD}_{\text{u}}^2 = & \frac{1}{|I_1|(|I_1|-1)} \sum_{i \in I_1} \sum_{j \in I_1, j \ne i} K(\X_i, \X_j) + \nonumber \\
& \frac{1}{|I_2|(|I_2|-1)} \sum_{i \in I_2} \sum_{j \in I_2, j \ne i} K(\X_i, \X_j) -\frac{2}{|I_1| \cdot |I_2|} \sum_{i \in I_1} \sum_{j \in I_2} K(\X_i, \X_j). \label{equ:ieq4}
\end{align}

Based on Eqn.~\eqref{equ:ieq3}\eqref{equ:ieq4} and Claim~7, $\text{\rm Pr}\left[\sup_{\X_1, \cdots \X_N \in \M} \left| \overline{\text{\rm MMD}}_{\text{b}}^2 - \text{\rm MMD}_{\text{b}}^2 \right| \ge \epsilon \right]$ is bounded by $C\left( \frac{\epsilon}{\sum_{(i,j)} |a_i a_j|} \right) = C\left(\frac{\epsilon}{4} \right)$, and then we obtain the uniform convergence for unbiased estimate of MMD.
\qed

\subsection*{A.4 Proof of Theorem~4}

Define $f(\X_i, \X_j; \W) = \Z'(\X_i; \W) \Z(\X_j; \W) - K(\X_i, \X_j)$ and recall that $\left| f(\X_i, \X_j) \right|\le 2$ and $\E \left[ f(\X_i, \X_j) \right] = 0$ (according to Lemma~3 in \citet{LeQ13}). Since $f$ is shift-invariant as their arguments, we use $\boldsymbol{\Delta} \equiv \X_i - \X_j \in \M_{\boldsymbol{\Delta}}$ for notational simplicity.

$\M_{\boldsymbol{\Delta}}$ is compact and with diameter at most twice $\operatorname{\text{\rm diam}}(\M)$. And then we can find an $\epsilon$-net that covers $\M_{\boldsymbol{\Delta}}$ using at most $T=\left( 4 \operatorname{\text{\rm diam}}(\M) / r \right)^d$ balls of radius $r$ \citep{Cucker01}. Let $\left\{\boldsymbol{\Delta}_k \right\}_{k=1}^T$ denote the centers of these balls.
We have $\left| f(\boldsymbol{\Delta}) \right| < \epsilon$ for all $\boldsymbol{\Delta} \in \M_{\boldsymbol{\Delta}}$ if the following two conditions hold for all $k$:
(i) $\left| f(\boldsymbol{\Delta}_k) \right| < \epsilon/2$;
(ii) $\left| f(\X_i, \X_j) \right| < \epsilon/2$, if $\X_i-\X_j$ belongs to the ball $k$ of the $\epsilon$-net.
Next we bound the probability of these two events:

(i) The union bound followed by Hoeffding's inequality applied to the anchors in the $\epsilon$-net gives
\begin{align}
\text{\rm Pr} \left[ \mathop{\cup}\nolimits_{k=1}^T | f(\boldsymbol{\Delta}_k) | \ge \epsilon/2 \right] \le 2T \exp\left( -d \epsilon^2/8 \right). \label{equ:ieq5}
\end{align}

(ii) If we use Fastfood for FastMMD in Algorithm~\ref{alg:alg2}, and suppose the estimate of kernel arises from a $d \times d$ block of Fastfood, then according to Theorem~6 in Fastfood literature \citep{LeQ13} we have:
\begin{align*}
& \text{\rm Pr} \left[ \left| f(\X_i, \X_j) \right|  \ge  \frac{2 \| \X_i - \X_j \|}{\sigma}  \sqrt{\frac{\log(2/\delta) \log(2d/\delta)}{d}} \right]
\le 2 \delta.
\end{align*}
Since $d$ usually is large in Fastfood method, we suppose that $(\log d) /2 \ge 1$. Considering Claim~8 and the fact $\| \X_i - \X_j \| \le 2 r$, this inequality can be further reformulated as
\begin{align}
& \text{\rm Pr} \left[ \left| f(\X_i, \X_j) \right| \ge  \epsilon \right]
\le  4 \exp\left( \frac{ -\sqrt{ (\log d)^2 + d \sigma^2 \epsilon^2/\| \X_i - \X_j \|^2} + \log d }{2} \right) \nonumber \\
& \le  4 \exp\left( \frac{ -\sqrt{ (\log d)^2 + 2^{-2} d \sigma^2 \epsilon^2/r^2} + \log d }{2} \right)
\le 64 \frac{\log d}{d \sigma^2 \epsilon^2} r^2. \label{equ:ieq6}
\end{align}

Combining \eqref{equ:ieq5} and \eqref{equ:ieq6} gives a bound in terms of the free variable $r$:
\begin{align*}
\text{\rm Pr} \left[\sup_{\X_1, \cdots, \X_N \in \M} \left| f(\X_i, \X_j) \right| \ge  \epsilon \right]
\le 2 \left( \frac{4 \operatorname{\text{\rm diam}}(\M)}{r} \right)^d \exp\left( \frac{-d \epsilon^2}{8} \right) + 64 \frac{\log d}{d \sigma^2 \epsilon^2} r^2.
\end{align*}
This has the form $1 - \kappa_1 r^{-d} - \kappa_2 r^2$. Setting $r = \left( \frac{\kappa_1}{\kappa_2} \right)^{\frac{1}{d+2}}$ turns this to $1 - \kappa_2^{\frac{d}{d+2}} \kappa_1^{\frac{2}{d+2}}$, and assuming that $ \frac{ \log d \operatorname{\text{\rm diam}}(\M)}{d \sigma^2 \epsilon^2} \ge 1$ and $\operatorname{\text{\rm diam}}(\M) \ge 1$, we obtain that
\begin{align}
\text{\rm Pr} \left[\sup_{\X_1, \cdots, \X_N \in \M} \left| f(\X_i, \X_j) \right| \ge  \epsilon \right]
\le 2^{12} \left( \frac{\log d \operatorname{\text{\rm diam}}(\M)}{d \sigma^2 \epsilon^2} \right)^2 \exp\left( -\frac{d \epsilon^2}{4(d+2)} \right) \triangleq C (\epsilon). \label{equ:ieq8}
\end{align}

Denote the right hand of the inequality~\eqref{equ:ieq8} as $C(\epsilon)$. Similar to the proof in Theorem~3, we have
\begin{equation}
\text{\rm Pr} \left[\sup_{\X_1, \cdots, \X_N \in \M} \left| \widehat{\text{\rm MMD}}_{\text{\rm b}}^2 - \text{\rm MMD}_{\text{\rm b}}^2 \right| \ge  \epsilon \right] \le C\left(\frac{\epsilon}{4}\right).
\end{equation}
Then we complete the proof of the uniform convergence for biased estimate of MMD. The convergence for unbiased estimate of MMD can be proved in the similar way.
\qed

\subsection*{A.5 Proof of Claim~5}

Let $p_1(x)$, $p_2(x)$ and $q(y)$ be the PDFs for random variables $X_1$, $X_2$ and $Y$, respectively. We then have
\begin{align}
& \E_{Q,P_1} \sin(Y - X_1) - \E_{Q,P_2} \sin(Y - X_2) \nonumber \\
=& \iint\! q(y)p_1(x) \sin(y-x) \, \mathrm{d}x \mathrm{d}y - \iint\! q(y)p_2(x) \sin(y-x) \, \mathrm{d}x \mathrm{d}y \nonumber \\
=& \int\! q(y) \left[ \int\! (p_1(x)-p_2(x)) \sin(y-x) \mathrm{d}x \right] \mathrm{d}y. \nonumber
\end{align}
Since $q(y) \ge 0$ and $\int\! q(y) \mathrm{d}y = 1$, the maximum of the previous expression with respect to all possible $q(y)$ is
\begin{equation}
\sup_y \int\! (p_1(x)-p_2(x)) \sin(y-x) \mathrm{d}x. \nonumber
\end{equation}
If $p_1(x)-p_2(x) = \sum_{i=1}^N a_i \delta(x-x_i)$, then
\begin{align*}
\int\! (p_1(x)-p_2(x)) \sin(y-x) \mathrm{d}x = \sum_{i=1}^N a_i \sin(y-x_i) = A \sin(y-\theta).
\end{align*}
The second equation holds because the sum of sinusoids with the same frequency is also a sinusoid with that frequency. According to the trigonometric identity, it holds that
\begin{align*}
A & = \left[ \sum_{i=1}^N \sum_{j=1}^N a_i a_j \cos(x_i-x_j) \right]^{\frac{1}{2}}, \\
\theta  & = \operatorname{atan2} \left( \sum_{i=1}^N a_i \sin x_i, \sum_{i=1}^N a_i \cos x_i \right),
\end{align*}
where $\operatorname{atan2}$ is the four-quadrant arctangent function. The supremum of this problem is $A$ when $y=\theta$.
\qed


\end{document}